\documentclass[journal]{vgtc}    

\usepackage{graphicx}
\graphicspath{{Figures/eps/}}
\DeclareGraphicsExtensions{.pdf,.png,.jpg,.eps}
\usepackage{times}                  
    
\usepackage{booktabs} 
\usepackage{lipsum}       
\usepackage{mwe}                      
\usepackage{newtxmath}
\usepackage{amsmath}
\usepackage{multirow} 
\usepackage{makecell}
\usepackage{booktabs}
\usepackage{colortbl}
\usepackage{tabularx}
\usepackage{mathtools}
\usepackage{textcomp}
\usepackage{siunitx}
\usepackage{algorithmic}
\usepackage{listings}
\usepackage{multirow}
\usepackage{boldline}
\usepackage{array}
\newcolumntype{C}[1]{>{\centering\arraybackslash}p{#1}}
\newcolumntype{M}[1]{>{\centering\arraybackslash}m{#1}}
\newcolumntype{L}[1]{>{\raggedright\arraybackslash}m{#1}} 
\usepackage{caption}
\usepackage[linesnumbered,ruled,lined,commentsnumbered]{algorithm2e}
\usepackage[percent]{overpic}
\usepackage[table]{xcolor}
\definecolor{lightgreen}{RGB}{200, 255, 200}
\definecolor{darkgreen}{RGB}{120, 200, 120}
\definecolor{darkgreen1}{RGB}{50, 120, 50}
\definecolor{lightred}{RGB}{255, 200, 200}
\definecolor{lightyellow}{RGB}{255, 255, 200}
\definecolor{lightyellow1}{RGB}{255, 220, 80}

\usepackage[labelformat=simple]{subcaption}

\usepackage{url}
\usepackage{ragged2e}
\usepackage{float}
\usepackage{pifont}

\onlineid{0}

\vgtccategory{Research}

\title{A Mixed Reality System for Robust Manikin Localization \\ in Childbirth Training}

\author{%
    \authororcid{Haojie Cheng}{0000-0002-9885-763X}, \authororcid{Chang Liu}{0009-0005-0235-8289}, \authororcid{Abhiram Kanneganti}{0000-0002-5559-4534}, \authororcid{Mahesh Arjandas Choolani}{0000-0001-8336-0973}, \\ \authororcid{Arundhati Tushar Gosavi}{0000-0002-8785-7335}, and \authororcid{Eng Tat Khoo$^{*}$}{0000-0003-1295-3506} 
}

\authorfooter{
  \item
  	Haojie Cheng, Chang Liu, Mahesh Arjandas Choolani, Arundhati Tushar Gosavi, and Eng Tat Khoo are with National University of Singapore.
  	E-mail: hjcheng@nus.edu.sg, chang\_lc@nus.edu.sg, obghead@nus.edu.sg, obggat@nus.edu.sg, etkhoo@nus.edu.sg.
   \item
    Abhiram Kanneganti is with National University Health System, Singapore.
   E-mail: abhiram\_kanneganti@nuhs.edu.sg.
   \item
    $^{*}$ Corresponding author.
}

\teaser{
\vspace{0.2in}
  \centering
  \subfloat[Pre-Delivery Phase]{%
    \includegraphics[height=0.245\linewidth]{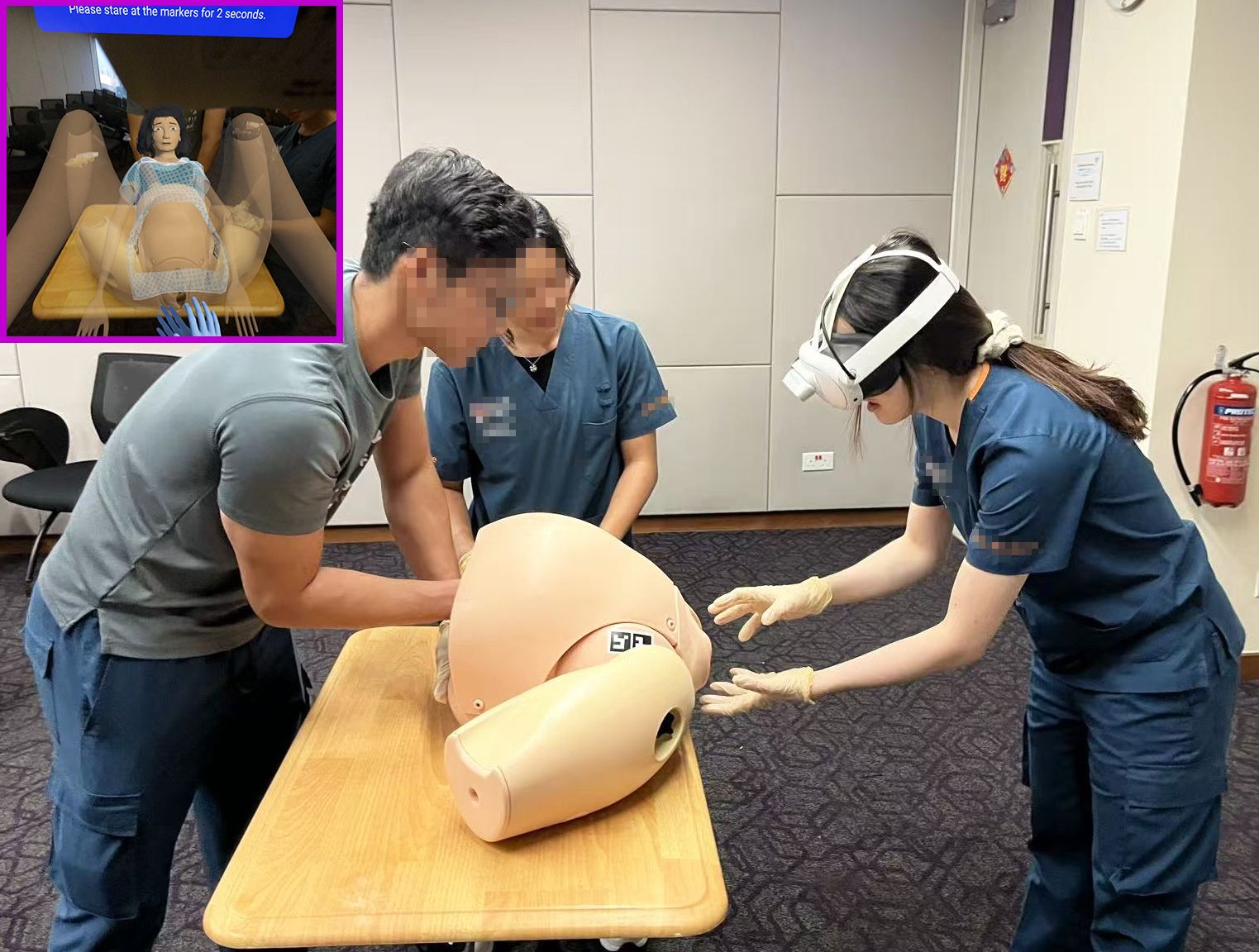}
  }
  \subfloat[Assisted-Delivery Phase]{%
    \includegraphics[height=0.245\linewidth]{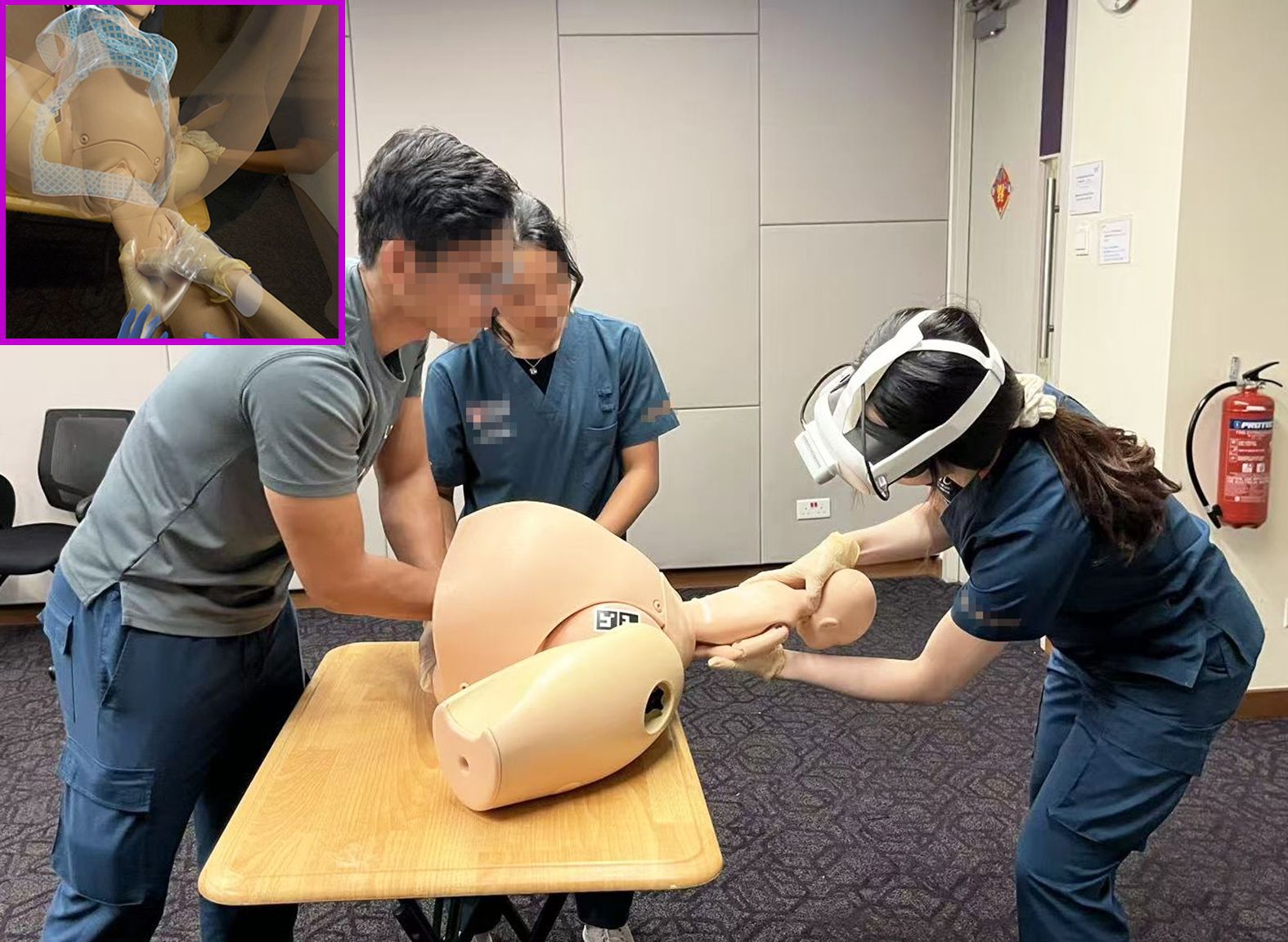}
  }
  \subfloat[Post-Delivery Phase]{%
    \includegraphics[height=0.245\linewidth]{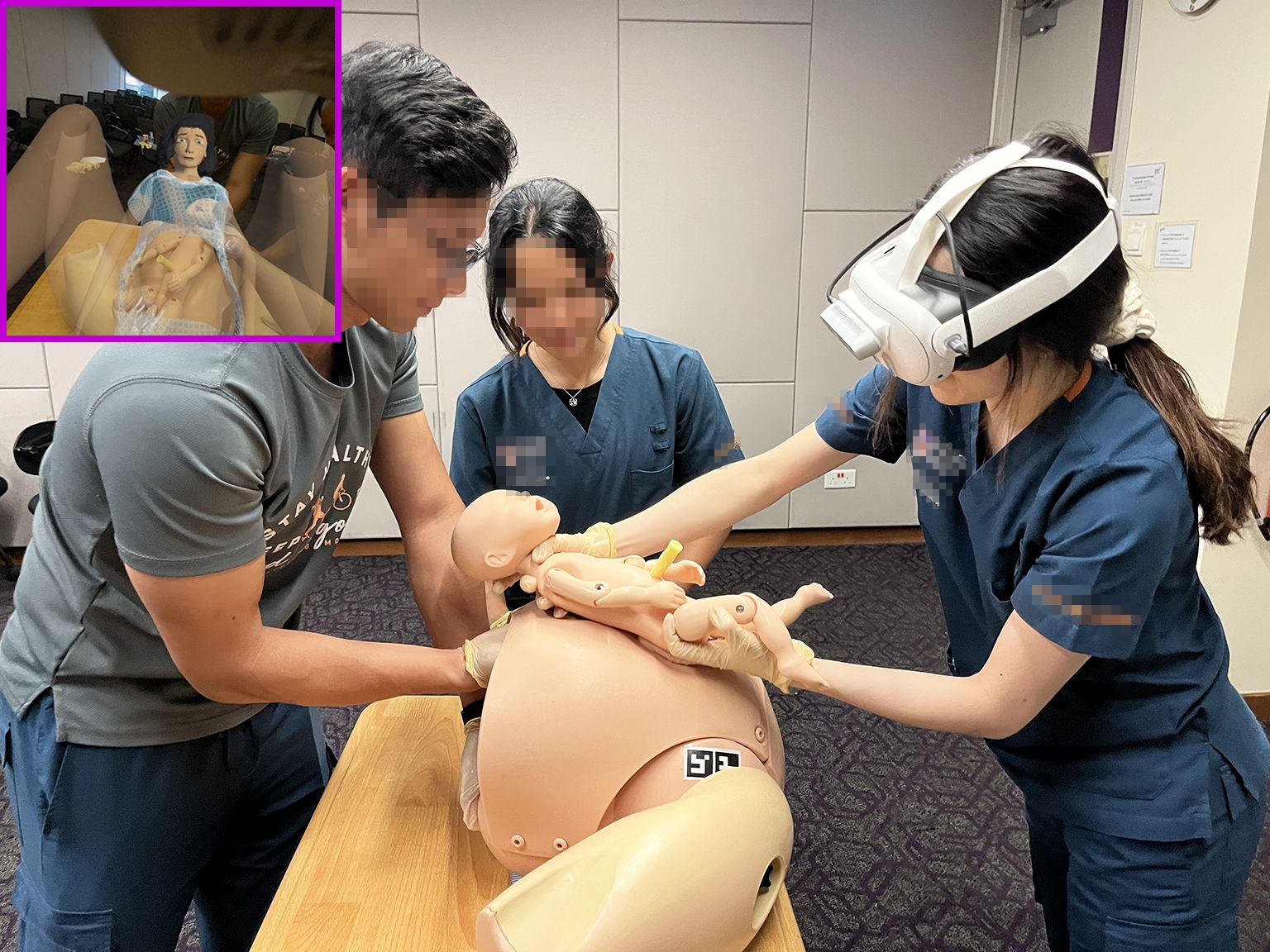}
  }
  \vspace{-0.05in}
  \caption{Mixed Reality-assisted childbirth training using a manikin setup. The primary trainee wears a Meta Quest headset with a front-facing RGB-D camera for real-time manikin localization. Two supporting trainees assist by pushing the neonatal manikin and stabilizing the manikin. The top-left image shows the trainee’s first-person view with virtual guidance.}
  \label{fig:teaser}
}
\abstract{
Opportunities for medical students to gain practical experience in vaginal births are increasingly constrained by shortened clinical rotations, patient reluctance, and the unpredictable nature of labour. To alleviate clinicians’ instructional burden and enhance trainees’ learning efficiency, we introduce a mixed reality (MR) system for childbirth training that combines virtual guidance with tactile manikin interaction, thereby preserving authentic haptic feedback while enabling independent practice without continuous on-site expert supervision. The system extends the passthrough capability of commercial head-mounted displays (HMDs) by spatially calibrating an external RGB-D camera, allowing real-time visual integration of physical training objects. Building on this capability, we implement a coarse-to-fine localization pipeline that first aligns the maternal manikin with fiducial markers to define a delivery region and then registers the pre-scanned neonatal head within this area. This process enables spatially accurate overlay of virtual guiding hands near the manikin, allowing trainees to follow expert trajectories reinforced by haptic interaction. Experimental evaluations demonstrate that the system achieves accurate and stable manikin localization on a standalone headset, ensuring practical deployment without external computing resources. A large-scale user study involving 83 fourth-year medical students was subsequently conducted to compare MR-based and virtual reality (VR)-based childbirth training. Four senior obstetricians independently assessed performance using standardized criteria. Results showed that MR training achieved significantly higher scores in delivery, post-delivery, and overall task performance, and was consistently preferred by trainees over VR training. Although validated only in the context of obstetric delivery, the system demonstrates strong potential for broader manikin-based procedural training and other healthcare education scenarios.
} % end of abstract


\begin{document}

\maketitle

\section{Introduction}\label{sec:introduction}

Witnessing the start of life in the labour ward is the hallmark of every medical student’s Obstetrics $\&$ Gynaecology (O$\&$G) rotation\cite{singh2005teaching}. In recent years, however, there has been a trend away from demonstrating competence in independently conducting routine vaginal births (VBs) towards one of participation or observation due to competing factors such as limited posting durations, patient aversion to male students \cite{akkad2008gender}. This reduction in opportunities for medical students to participate in VBs is worrying considering that non-obstetricians and -midwives are frequently first responders to unexpectedly precipitous VBs occurring out-of-hospital or in acute care settings \cite{Dotters2018Preclinical, Morgan2014Preparing} and that obstetrics comprises a large and continually rising proportion of medicolegal liabilities in developed countries \cite{wilson2016simulation, jones2024royal}. Traditional group-based teaching methods include a combination of didactic lectures, manikin simulation sessions, and delivery suite tours \cite{Daniels2010Prospective,Birch2007Obstetric,Sutherland2012Teaching}. While lectures and materials can be standardized, practical exposure to childbirth remains limited and unpredictable. These constraints underscore the need for more efficient training methods that ensure students gain both theoretical knowledge and hands-on experience in childbirth procedures.

% Many students miss the opportunity to observe live births during clinical rotations due to the spontaneous nature of labor, reduced clerkship durations, and instructors' clinical workload \cite{Dotters2018Preclinical, Morgan2014Preparing}.

The growing adoption of immersive technologies, particularly virtual reality (VR) and mixed reality (MR), is reshaping medical education by offering scalable, interactive training solutions \cite{barrie2019mixed, turso2023virtual, ng2022real}. In childbirth training, where clinical exposure is often limited, these technologies present valuable alternatives. VR enables the simulation of fully virtual delivery suites with interactive guidance for childbirth procedures \cite{chang2024effects}. Previous research \cite{liu2024facilitating} demonstrated that students who trained with a VR-based childbirth simulator achieved significantly greater knowledge gains compared to those who underwent traditional manikin-based simulation. However, a key limitation of VR training lies in its fully virtual environment, which precludes interaction with the physical surroundings and lacks tactile feedback. Other studies have also shown that tactile feedback is crucial in learning procedural skills \cite{benjamin2021using,sigrist2013augmented,cheung2023medical,yang2020immersive, rangarajan2020systematic}. To bridge this gap, MR offers a hybrid approach that combines virtual guidance with physical interaction \cite{ljungblad2025mixed,sielhorst2004augmented,cheng2023realistic, cheng2023mixed}. In childbirth training, MR allows students to perform procedures on physical manikins while receiving real-time visual cues, such as hand manoeuvre guidance. This integration retains the tactile realism of manikin-based training while enhancing it with interactive and adaptive instruction.

%Figure 
\begin{figure*}[t]
% \vspace{-0.10in}
\centering
\includegraphics[width=1.0\textwidth]{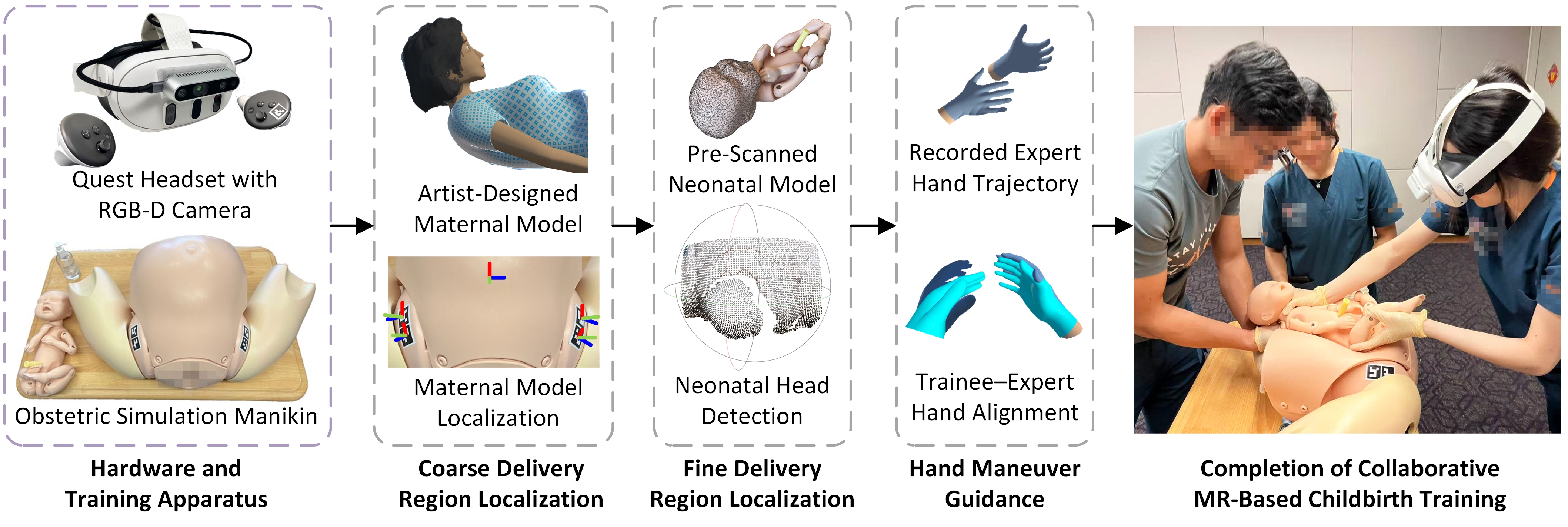}
\vspace{-0.20in}
\caption{Workflow of our MR visualization framework for immersive childbirth training, employing a coarse-to-fine delivery region localization strategy that sequentially estimates maternal and neonatal poses to ensure accurate hand maneuver guidance for collaborative training.}
\label{fig:Pipeline}
% \vspace{-0.10in}
\end{figure*}

However, integrating MR into childbirth training for large-scale medical education presents several technical and practical challenges. A key difficulty lies in accurately localizing physical manikins within the MR environment. During delivery simulation, expulsion forces and abundant lubrication, coupled with the non-rigid structure and silicone material of the manikin, which makes robust localization particularly difficult. To be suitable for widespread use in medical education settings, the MR system must be simple to set up and consistently deployable, enabling large cohorts of students to engage with the system within their limited rotation times. It must also be autonomous and easy to operate, allowing medical students with no technical background to use the system independently without the assistance of specialized personnel or the need for complex calibration procedures.

To address these challenges, we present the first MR system for childbirth training, targeting normal delivery as the most common and fundamental clinical scenario. The system emphasizes essential competencies in the correct execution of delivery maneuvers, confident post-delivery handling, and accurate neonatal placement on the maternal abdomen. To enable spatial alignment between virtual and physical models, an external RGB-D camera is integrated with the headset and calibrated through an improved eye-to-hand method. Accurate localization is then achieved through a coarse-to-fine registration pipeline: the virtual maternal model is first aligned to the manikin using fiducial markers to define a structurally correct delivery region of interest (ROI), and the pre-scanned neonatal head is subsequently registered within this ROI. This process allows virtual guiding hands to be overlaid on the manikin with anatomical accuracy, enabling trainees to follow expert trajectories reinforced by haptic feedback. A large-scale user study was conducted with 83 fourth-year medical students, whose performance was evaluated by four senior obstetricians. The results demonstrated that MR training led to greater engagement and significantly higher scores in delivery, post-delivery, and overall handling tasks compared to VR training.

The main contributions of our paper are outlined as follows:
% \vspace{-0.03in}
\begin{itemize}
\item We propose the first MR-based childbirth training system that integrates virtual guidance with tactile manikin interaction, allowing trainees to practice independently without on-site expert supervision.\vspace{0.05in}

\item We present the first integration of an external RGB-D camera with a Quest headset in passthrough mode, enabling real-time communication and stable MR interaction.\vspace{0.05in}

\item We develop a coarse-to-fine localization strategy combining multi-marker alignment of the maternal manikin with point cloud registration of neonatal head to compute accurate virtual guiding hand positions.\vspace{0.05in}

\item We conducted a large-scale MR user study for childbirth training, which to our knowledge provides initial empirical insights into how MR-based training differs from VR-based approaches in this domain.
\end{itemize}

% \vspace{-0.10in}
\section{Related Research}
\label{sec:related_work}

\subsection{Limitations of Conventional Childbirth Training} 

Childbirth training for medical and midwifery students relies heavily on labour ward placements, but opportunities to observe or assist in deliveries are often limited by unpredictable clinical schedules and the availability of consenting patients \cite{cotter2016medical, grasby2001attitudes}. The spontaneous and time-sensitive nature of labour further restricts access to meaningful learning experiences \cite{chang2010effect, mavis2006medical}.

%  Ethical considerations, such as the intimate setting of childbirth and patient hesitancy, particularly toward male students, also reduce participation \cite{chang2010effect}. These challenges are compounded by shortened clinical rotations and increasing demands on educators \cite{mavis2006medical}.

To compensate for limited clinical exposure, conventional childbirth education often relies on slideshows, instructional videos, and manikin-based simulations. Slideshow-based teaching remains common in preclinical settings due to its simplicity, but it does not adequately prepare students for the dynamic and tactile nature of childbirth \cite{Dotters2018Preclinical}. Instructional videos help familiarize students with the delivery environment and improve engagement \cite{Chang2019}, yet they are passive and do not support skill development through active practice \cite{Birch2007Obstetric}. Physical manikins are widely used to simulate childbirth and obstetric emergencies \cite{Daniels2010Prospective}, offering valuable procedural practice. However, they depend on the presence of experienced instructors to guide learners and provide timely feedback, which limits scalability and consistency across large student groups \cite{Sutherland2012Teaching}. These limitations highlight the need for more interactive and scalable training systems that provide realistic procedural experience alongside individualized support.

\subsection{VR and MR Applications in Childbirth Training}

To address the limitations of conventional methods, immersive technologies such as VR and MR have been increasingly adopted to enhance procedural training and learner engagement in childbirth training. Chang et al. \cite{Yao2019} developed the IFOREAL simulator with 3D anatomical animations and delivery tools, improving perceived usefulness but hindered by unfamiliar controls. To further enhance intuitiveness, Nugraha et al. integrated Leap Motion-based hand tracking for natural hand interactions \cite{Nugraha2018}. More recently, Liu et al. developed a comprehensive VR childbirth simulator with guided hand manoeuvres and conducted a large-scale empirical study with medical students \cite{liu2024facilitating}. While the system significantly improved learning outcomes compared to manikin training, students reported lower ratings in usability and feedback, reflecting VR’s ongoing limitations in tactile realism and instructor presence.

% Despite this advancement, midwives in their study struggled to complete the training independently, underscoring the need for greater accessibility and usability in VR systems.

To bridge the gap between virtual guidance and physical interaction, several MR systems have been explored for childbirth training. Mazalan et al. \cite{Mazalan2023} proposed a marker-based MR application using Vuforia to display 3D childbirth animations on smartphones with interactive quizzes, but its limited immersion and lack of tactile realism positioned it more as a reference tool than a simulator. Building on interactivity, Ballit et al. \cite{Ballit2023Fast} developed an MR simulation on HoloLens 2 that applied a hyperelastic mass-spring model to reproduce fetal and pelvic tissue deformation, enabling pseudo-haptic practice of forceps delivery. Yet, in the absence of user studies or clinical context, its educational efficacy remains uncertain. Complementarily, Ljungblad et al. \cite{ljungblad2025mixed} employed MagicLeap 2 for MR anatomical visualization, allowing learners to inspect a virtual fetus within the maternal pelvis. While well received for conceptual understanding, it provided little support for procedural training or psychomotor skill acquisition. Collectively, these efforts underscore that existing MR systems tend to emphasize either accessibility or visualization, but fall short of delivering immersive, validated, hands-on childbirth training.

\subsection{Vision-based Manikin Tracking Techniques in MR} 

Accurate tracking of physical manikins is essential for immersive interaction in MR-based childbirth simulations. Marker-based tracking is a mature and widely adopted technique in medical MR systems due to its high spatial accuracy, low computational overhead, and robustness on mobile HMDs. One common approach employs planar fiducial markers to register patient anatomy or instruments with virtual overlays to assist with navigation tasks such as ventriculostomy and neuronavigation \cite{schneider2021augmented, frantz2018augmenting}. To enhance robustness and minimize drift, several studies have introduced multi-marker fusion or fixed marker rigs, which improve tracking consistency under head or patient movement \cite{fick2021holographic, lai2020fusion}. Some systems further integrate marker detection with the optical pipeline of HoloLens or other wearable devices, enabling direct marker recognition without external cameras \cite{eom2022neurolens, zhang2020research, ng2023holopocus}. Building on these insights, effective deployment requires careful planning of marker placement to ensure visibility and stability during interaction, especially in constrained clinical or training environments.

To reduce reliance on physical markers, deep learning-based 6-DoF object pose estimation has gained traction in recent years. Methods such as FoundationPose \cite{wen2024foundationpose}, MegaPose \cite{labbe2022megapose}, NeRF-Pose \cite{li2023nerf}, Any-6D \cite{lee2025any6d} and OSOP \cite{shugurov2022osop} infer object poses directly from RGB or RGB-D inputs. These models are primarily designed for rigid objects, which makes them difficult to adapt to deformable training manikins used in childbirth education. To address this limitation, human pose estimation techniques have been explored for tracking flexible manikins \cite{cao2017realtime, jiang2024manikin}. Some studies further extend these capabilities to specialized scenarios, such as infant pose estimation \cite{groos2022towards, kyrollos2023under, scherfgen2021estimating}. However, they remain highly sensitive to occlusions, particularly when the trainee’s hands or instruments cover parts of the manikin. In addition, their dependence on high-performance GPU inference makes real-time deployment on mobile HMDs infeasible without incurring significant latency or requiring offloading to external servers.

% These methods aim to capture articulated motion without relying on predefined object geometry.

\section{Methodology}
\label{sec:overview}

\subsection{Design of Our MR Childbirth Training System}
\label{subsec:SystemSetup}

Our work aims to integrate the benefits of virtual feedback and physical manikin training to develop an interactive MR childbirth training system. Even in the absence of professional obstetricians, our MR system can assist medical students and junior doctors in familiarizing themselves with and performing the correct procedures during childbirth, as shown in Fig. \ref{fig:Pipeline}.

Compared to the HoloLens headset commonly used in MR tasks\cite{eom2022neurolens, cheng2023realistic, ng2023holovein}, the Quest headset demonstrates superior performance in key aspects such as screen field of view, resolution, refresh rate and processor performance \cite{ungureanu2020hololens}. Furthermore, the Quest headset supports seamless switching between VR and MR modes. This flexibility allows direct comparisons between VR and MR experiences in our user study under identical hardware conditions, which is not feasible with the HoloLens headset due to its lack of VR support. Although recent firmware updates provide limited access to the Quest’s camera, the interface suffers from low transmission speed and the absence of spatially aligned RGB-D streams, making it unsuitable for real-time MR interaction. To address this limitation, our system first establishes a customized data access pipeline and integrates an external RGB-D camera onto the Quest headset, enabling reliable, spatially aligned real-world scene capture in real time to track the $3$D pose of the maternal and neonatal manikins.

A central requirement for effective normal delivery training is that trainees learn the correct sequence of hand maneuvers. To support this goal, our system must first localize the delivery region of the physical obstetric manikin with sufficient accuracy to guide procedural interaction. We adopt a coarse-to-fine localization strategy: the maternal manikin is first used to define a coarse delivery region, and the neonatal manikin’s head position is then employed to refine this region to a clinically meaningful scale. To meet physicians’ requirements for large-scale, user-friendly deployment, the entire childbirth training system operates on a standalone headset without the need for external servers. For maternal manikin tracking, we employ a fiducial marker–based approach, which provides efficient and reliable registration even when no exact virtual replica of the maternal model is available. In contrast, the neonatal manikin is composed entirely of silicone and undergoes significant deformation during simulated delivery, making marker attachment infeasible. To address this challenge, we implement a point cloud registration method on the neonatal head, which refines the delivery region dynamically. Once the delivery region is established, the system overlays virtual guiding hands in the corresponding spatial location. Trainees are instructed to follow the guiding trajectory and practise the correct hand maneuvers on the physical manikin, thereby combining tactile realism with real-time procedural guidance in MR.

\subsection{Multi-Coordinate System Integration}
\label{subsec:CoordinateIntegration}

Accurate spatial registration is fundamental to enabling realistic interaction in MR-based childbirth training. Since the Quest headset, the mounted RGB-D camera, and the obstetric manikin each operate in separate local coordinate systems, a consistent unification is required to ensure that virtual overlays align precisely with the physical manikin. All components are integrated into the headset coordinate system, which serves as the global reference for both virtual rendering and physical interaction.

To localize the maternal manikin in the headset frame, we first adopt the headset coordinate frame $H$ as the global reference, since rendering and user input are all anchored to the HMD, expressing all poses in $H$ avoids downstream frame conversions. We then introduce two time-invariant links to reduce the estimation degrees of freedom: the RGB-D camera $C$ is rigidly mounted on the headset, yielding a constant transform $\mathbf{T}^H_C$, and fiducial markers $F$ are rigidly attached to the maternal manikin $M$ at a known location, yielding a constant $\mathbf{T}^F_M$. At runtime, the reliably observable quantity is the marker pose in the camera frame, $\mathbf{T}^C_F(t)$, obtained via fiducial detection. By composing transforms along the kinematic chain from $M$ to $H$, the manikin pose in the headset frame follows as:

\vspace{-0.05in}
\begin{equation}
\mathbf{T}^H_M(t) = \mathbf{T}^H_C \cdot \mathbf{T}^C_F(t) \cdot \mathbf{T}^F_M .
\label{eq:manikin_in_headset}
\end{equation}

Consequently, once $\mathbf{T}^C_F(t)$ is estimated, $\mathbf{T}^H_M(t)$ is uniquely determined by Eq. \ref{eq:manikin_in_headset}. Fusion process across multiple visible markers and the calibration of $\mathbf{T}^H_C$ are detailed in Secs.~\ref{subsubsec:MotherTracking} and~\ref{subsec:QuestRGBD}, respectively.

%Figure 
\begin{figure}[h]
  \vspace{-0.05in}
\centering
\includegraphics[width=0.43\textwidth]{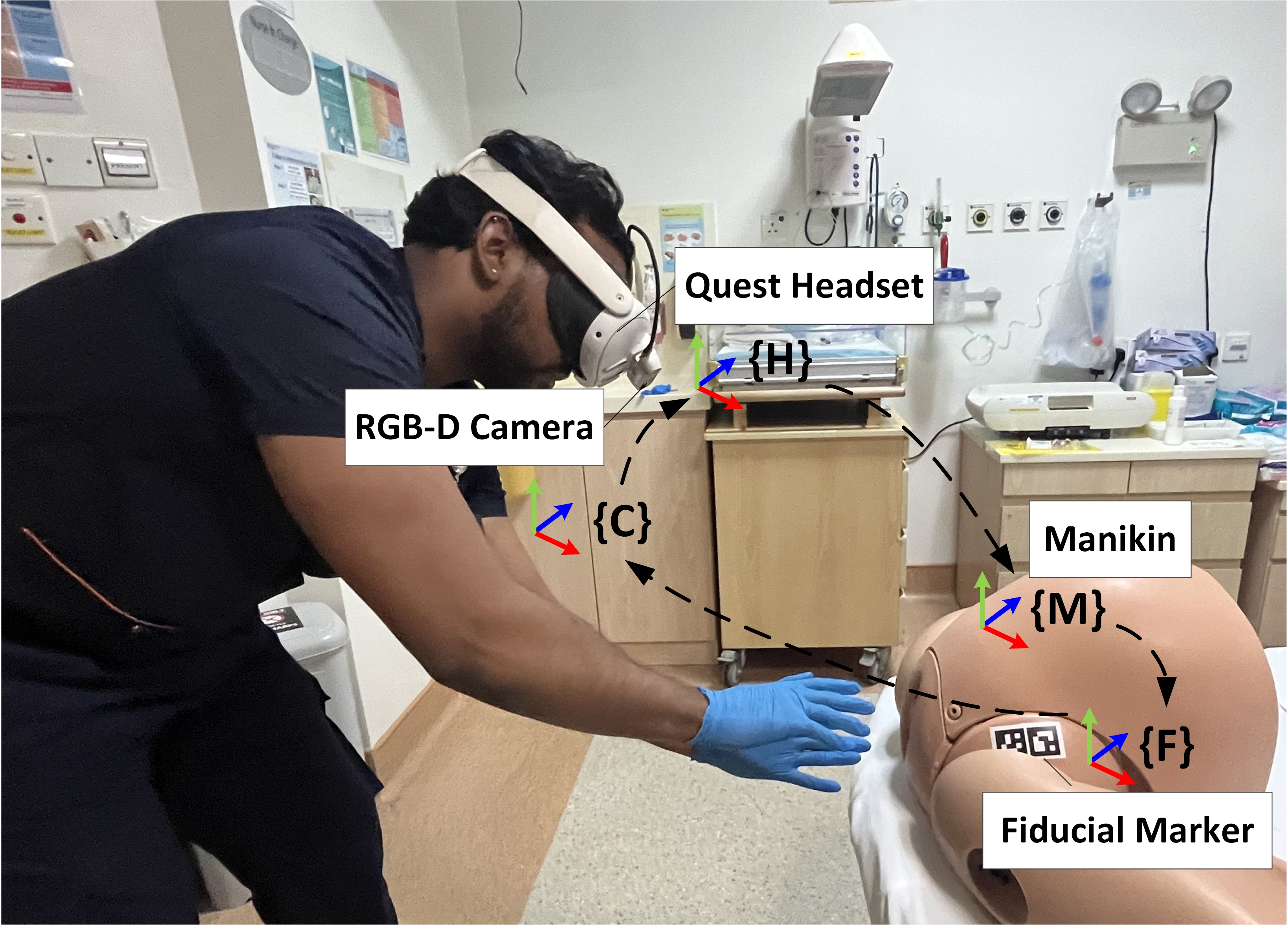}
\caption{System setup and transformation of world coordinate systems between Quest headset and manikin.}
\label{fig:coorsystem}
\vspace{-0.10in}
\end{figure}

\subsection{Spatial Registration of Headset and RGB-D Camera}
\label{subsec:QuestRGBD}

Since the Quest headset and the RGB-D camera occupy different spatial positions and operate in distinct coordinate frames, camera calibration is required to ensure precise virtual-to-physical alignment in MR display. Conventional calibration methods typically rely on capturing fiducial markers simultaneously across cameras and solving for their relative transformation \cite{arucomarker2014}. However, the Quest interface does not provide access to reliable RGB-D streams or intrinsic parameters of its onboard cameras, which makes direct cross-camera calibration with the external RGB-D device infeasible. Consequently, we estimate the relative transformation between the headset and the RGB-D camera through a custom calibration procedure described below.

Since the Quest headset provides access to the controller's pose through its built-in development modules. Meanwhile, the RGB-D camera can capture images containing fiducial markers. Building on this setup, we pioneer the application of dynamic pose tracking using the Quest controller and fiducial markers to extend the eye-to-hand calibration algorithm \cite{shiu1987calibration}. The RGB-D camera and fiducial marker should be rigidly attached to the Quest headset and Quest controller respectively, to maintain the fixed relative pose. According to the rigid body kinematics theory \cite{shiu1987calibration}, the coordinate system transformation within the setup between two adjacent frames $t_i$ and $t_j$ can be described as: 

% Equation 1
\begin{equation}\label{eq:eye2hand1}
% \vspace{-0.10in}
\mathbf{T}^{U}_{H}(t_i, t_j) \cdot \mathbf{T}^{H}_{C} = \mathbf{T}^{H}_{C} \cdot \mathbf{T}^{C}_{F}(t_i, t_j),
% \vspace{-0.05in}
\end{equation}

{\noindent}where $\mathbf{T}^{U}_{H}(t_i, t_j) = \mathbf{T}^{U}_{H}(t_j) \cdot (\mathbf{T}^{U}_{H}(t_i))^{-1}$ and $\mathbf{T}^{C}_{F}(t_i, t_j) = \mathbf{T}^{C}_{F}(t_j) \cdot (\mathbf{T}^{C}_{F}(t_i))^{-1}$. Specifically, $\mathbf{T}^{U}_{H}(t_i, t_j) \in \mathbb{R}^{4\times 4}$ is the relative transformation of the user controller $U$ with respect to headset $H$, and $\mathbf{T}^{R}_{M}(t_i, t_j)\in \mathbb{R}^{4\times 4}$ is the relative transformation of RGB-D camera $C$ with respect to the fiducial marker $F$. The relative pose between $H$ and $C$ can be subsequently calculated as:

% Equation 2
\vspace{-0.10in}
\begin{equation}\label{eq:eye2hand2}
% \vspace{-0.10in}
\mathbf{T}^{H}_{C} = \underset{\mathbf{T}^{H}_{C}}{\text{argmin}} \sum_{i,j} \| \mathbf{T}^{U}_{H}(t_i, t_j) \cdot \mathbf{T}^{H}_{C} - \mathbf{T}^{H}_{C} \cdot \mathbf{T}^{C}_{F}(t_i, t_j) \|.
\vspace{-0.05in}
\end{equation}

This constrained nonlinear homogeneous equation can be solved by Tsai-Lenz algorithm \cite{tsai1989new}. In practical applications, the fiducial marker can be affixed to the Quest controller at any position to ensure a rigid connection. After completing the precomputed camera calibration, the real-world scene information captured by the RGB-D camera can be accurately aligned and rendered in the passthrough mode of Quest headset, ensuring correct spatial positioning and seamless integration.

\vspace{0.10in}
\subsection{Design of Obstetric Manikin Tracking}
\label{subsec:ManikinTracking}

To enable procedural guidance and assessment on a standalone headset, our tracker delivers stable, low-latency manikin poses within a coarse-to-fine pipeline that anchors the delivery region from the maternal reference and refines it using neonatal head pose for precise virtual guidance.

\subsubsection{Fiducial Marker-based Maternal Manikin Tracking}
\label{subsubsec:MotherTracking}

The fiducial marker offers an optimal trade-off between tool cost, implementation feasibility and tracking stability. However, fiducial markers on the curved manikin surface often produce temporally unstable 3D pose estimations with RGB-only detection methods \cite{wieczorek2020effects}. To address this, we optimized the detection process by adding depth information. Specifically, the RGB image is used to detect the 2D positions of the fiducial marker corners, while the camera's intrinsic parameters for both RGB and depth sensors are utilized to calculate the 3D positions of the marker corners. These 3D positions are then integrated to derive more robust and accurate pose parameters of the fiducial marker.

% \vspace{-0.10in}
\begin{figure}[h]
\centering
\subfloat[Unlocalized Manikin]{
  \hspace{-0.05in}
\includegraphics[height=0.17\textwidth]{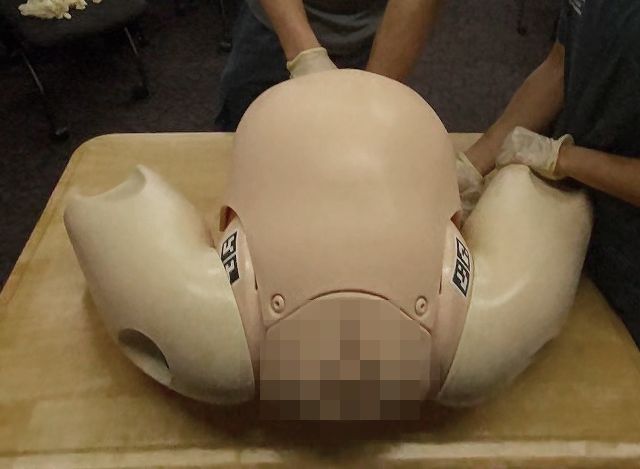}
}
\subfloat[Localized Manikin with Virtual Overlay]{
  % \hspace{-0.05in}
\includegraphics[height=0.17\textwidth]{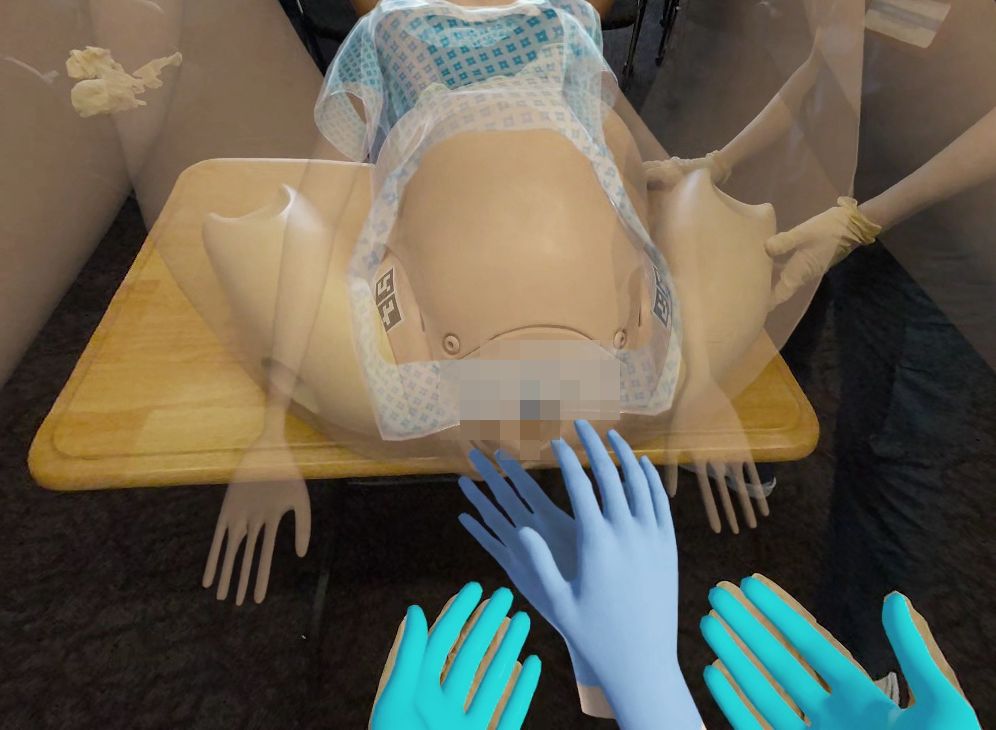}
}
\vspace{-0.10in}
\caption{First-person headset view of manikin localization. The light-blue hands indicate the tracked hands of the trainee, while the dark-blue hands represent the expert’s virtual guiding hands.}
\label{fig:mothermodel}
% \vspace{-0.10in}
\end{figure}

To robustly achieve accurate maternal model alignment in MR, multiple fiducial markers are strategically placed on the maternal manikin's surface (Fig.~\ref{fig:mothermodel}) to expand viewpoint coverage and reduce the impact of single-marker occlusion. After defining the marker coordinate system, the relative poses of all markers with respect to the virtual maternal model’s center are precomputed during initialization, thereby establishing a fixed geometric prior and avoiding per-session recalibration. The fixed spatial relationship between the $i$-th marker $F_i$ and the model is denoted by $\mathbf{T}_{F_i}^{M}$ so that each detected marker yields a consistent candidate estimate of the model pose.

During runtime, the headset-mounted RGB-D camera continuously detects the live poses of visible markers. Given each marker pose $\mathbf{T}_{F_i}$ in the RGB-D camera coordinate system, the associated local model pose $\mathbf{T}^{F_i}_{M}$ is obtained by chaining the measurement with the precomputed marker-to-model transform. When multiple markers are visible during training, their per-marker estimates are fused to determine the optimal model pose $\mathbf{T}_{M}$, leveraging redundancy to mitigate measurement noise and transient occlusions. A distance-weighted scheme is adopted, with $w^i$ inversely proportional to the squared camera–marker distance, so that closer observations contribute more to the fused estimate:

\vspace{-0.05in}
\begin{equation}\label{eq:optimal_vm_rgbd}
\mathbf{T}^{F}_{M} = \mathcal{F}(\sum_{i=1}^N \tilde{w}^i \cdot \mathbf{T}^{F_i}_{M}), \quad \tilde{w}^i = \frac{w^i}{\sum_{j=1}^N w^j},
\end{equation}

\noindent where $\mathcal{F}(\cdot)$ denotes the one Euro filter~\cite{casiez20121}, which suppresses high-frequency jitter while preserving responsiveness required for interactive guidance. After localizing the physical maternal manikin, the preset local pose of the virtual guiding hands $\mathbf{G}$ relative to the virtual maternal model and the calibrated transform $\mathbf{T}^{H}_{C}$ in Eq.~(\ref{eq:eye2hand2}) are incorporated to express guidance in the headset frame:

% \vspace{-0.05in}
\begin{equation}\label{eq:optimal_vm_quest}
\mathbf{T}^{H}_{\text{hand}} =\mathbf{G} \cdot \mathbf{T}^{H}_{C} \cdot \mathbf{T}^{C}_{F}\cdot \mathbf{T}^{F}_{M}.
\end{equation}

In our MR system, the $3$D model of the maternal manikin is designed by digital artists and does not perfectly replicate the shape or dimensions of the physical manikin. As a result, the estimated hand positions derived from Eq. (\ref{eq:optimal_vm_quest}) can only serve as coarse approximations. Therefore, upon determining the initial coarse hand pose $\mathbf{T}^{H}_{\text{hand}} \in \mathbb{R}^{4\times 4}$, a delivery ROI is instantiated based on the physical maternal manikin’s dimensions, thereby constraining subsequent neonatal head localization and reducing computational complexity.

\subsubsection{Markerless Neonatal Manikin Detection}
\label{subsubsec:BabyDetection}

To refine and update the virtual guiding hand placement based on the actual position of the neonatal manikin, it is necessary to perform a secondary localization step once the neonate begins to emerge from the vaginal opening. Since normal delivery tasks do not require precise or continuous tracking of the neonatal pose, this localization only needs to be performed once, and can be accomplished without relying on computationally intensive data-driven methods or additional external computing devices.

\begin{figure}[h]
\centering
\subfloat[RGB and Depth Image Pair]{
  \hspace{-0.05in}
\includegraphics[height=0.3\textwidth]{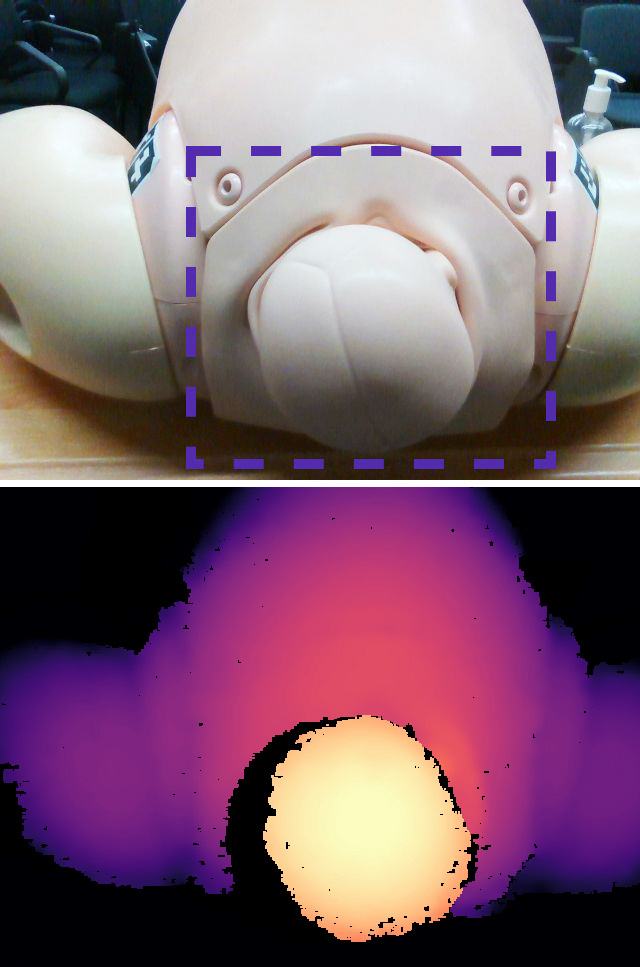}
}  \hspace{0.05in}
\subfloat[Point Cloud and Normal of ROI]{
\includegraphics[height=0.3\textwidth]{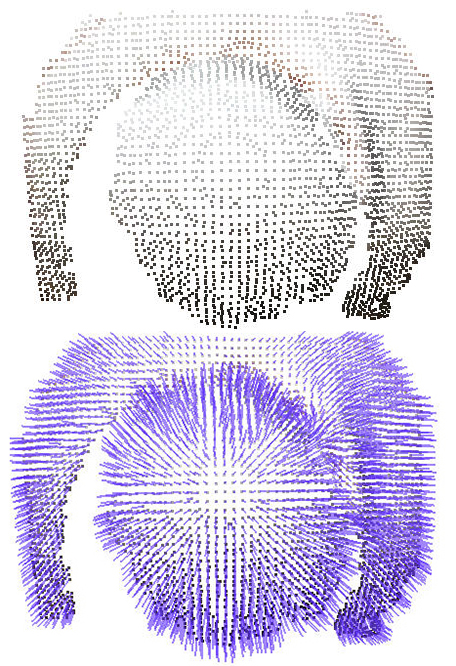}
}
\vspace{-0.10in}
\caption{Point cloud generation of the neonatal head ROI. The purple dashed box in the RGB image indicates a coarse search region near the delivery area for localizing the neonatal head.}
\label{fig:rgbddetect}
% \vspace{-0.10in}
\end{figure}

Although fiducial markers are widely used for tracking physical objects because of their simplicity and reliability, they are not well suited to a neonatal manikin in childbirth training. The manikin’s silicone surface lacks sufficiently planar regions, which prevents secure adhesion of markers. During simulated delivery, the head and body also experience substantial compression and shear that further destabilize any attachment. In addition, the lubricant applied to facilitate delivery produces specular reflections that interfere with detection and gradually degrades the printed markers. Taken together, these factors make marker-based tracking unreliable for the neonatal manikin.

% \vspace{-0.05in}
\begin{figure}[h]
\centering
\hspace*{-0.10in}
\subfloat[Digitized Neonatal Model]{
  \includegraphics[height=0.14\textwidth]{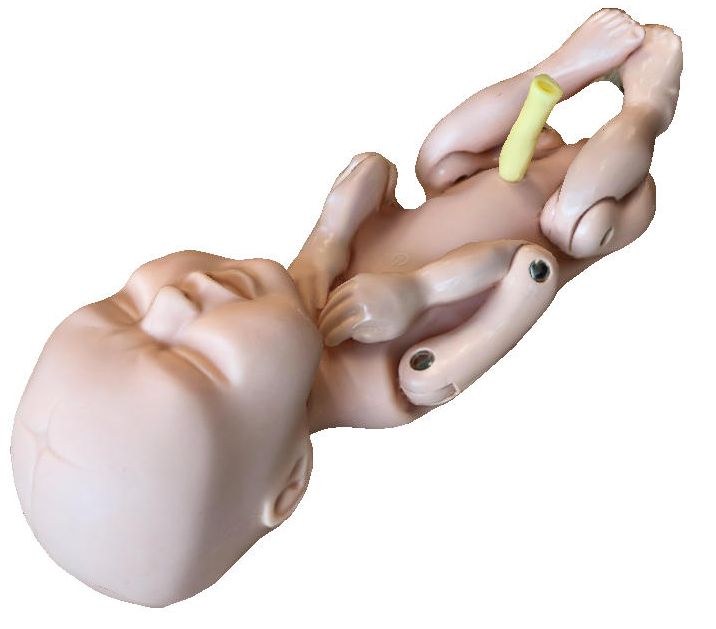}
}
 \hspace{0.02in}
\subfloat[Neonatal Head Mesh]{
\includegraphics[height=0.14\textwidth]{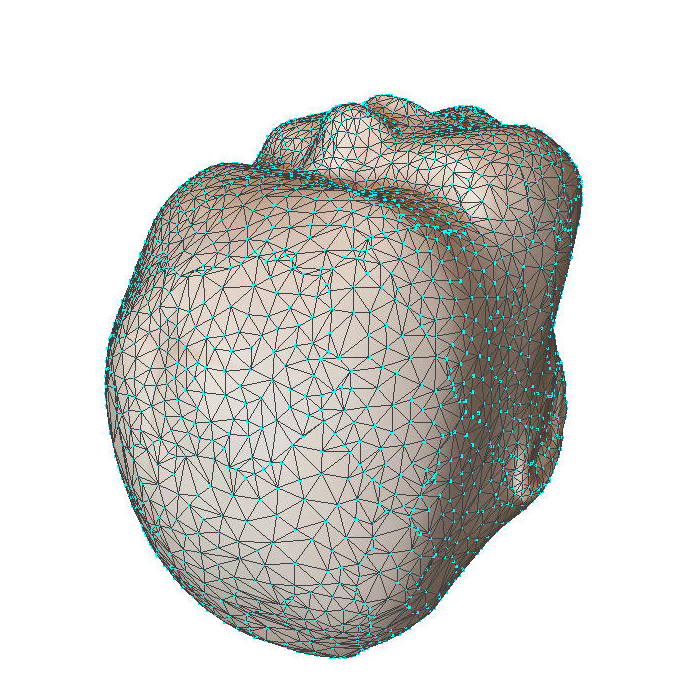}}
\hspace{0.02in}
\subfloat[Point Cloud Registration]{
\includegraphics[height=0.135\textwidth]{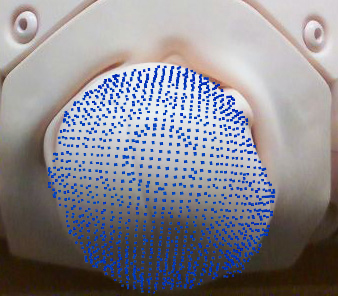}}
\vspace{-0.10in}
\caption{Virtual neonatal head model acquisition and point cloud registration for localization in an MR environment.}
\label{fig:babypoint}
\vspace{-0.10in}
\end{figure}

To robustly detect the initial emergence of the neonatal manikin, we estimate the head pose using a fast two-stage point-cloud registration pipeline. A high-resolution template point cloud of the neonatal head $\mathbb{P}$ is preacquired with a structured-light scanner and registered offline to the MR system’s global frame. At runtime, the current coarse hand pose $\mathbf{T}^{H}_{\text{hand}}$ (Eq.~\ref{eq:optimal_vm_quest}) defines a delivery ROI near the outlet (Fig.~\ref{fig:rgbddetect}), which bounds the search to the clinically relevant area, reduces computation, and suppresses spurious correspondences from background geometry. Within this ROI, we adopt Fast Global Registration (FGR) \cite{zhou2016fast} for coarse initialization because the neonatal head frequently enters with unknown orientation and partial visibility. FGR estimates a global rigid transform directly from feature correspondences without an initial guess, satisfying our time-constrained, no-prior requirement and furnishing a suitable initializer for local refinement. The coarse alignment is then refined with point-to-plane ICP \cite{chen1992object}, which is well matched to the relatively rigid head geometry in the localized ROI, offering fast convergence and improved metric accuracy under small residual misalignments, thereby meeting the latency budget on a standalone headset. The resulting neonatal head pose $\mathbf{T}_{b}$ (Fig.~\ref{fig:babypoint}) is used to update the relative trainee-hand transform $\hat{\mathbf{G}}$ in conjunction with the current maternal model pose $\mathbf{T}^{F}_{M}$ (Eq.~\ref{eq:optimal_vm_rgbd}), re-anchoring virtual guidance to the observed contact geometry. The updated hand pose in the headset coordinate system is then computed by the same transform chaining as in Sec.~\ref{subsec:ManikinTracking}:

% Equation 7
% \vspace{-0.05in}
\begin{equation}\label{eq:final_hands_pose}
\hat{\mathbf{T}}^{H}_{\text{hand}} =\hat{\mathbf{G}} \cdot \mathbf{T}^{H}_{C} \cdot \mathbf{T}^{C}_{F}\cdot \mathbf{T}^{F}_{M}.
\end{equation}

% \vspace{-0.05in}
\subsection{MR-Guided Hand Manoeuvres for Normal Delivery}
\label{subsec:UIdesign}

To enable effective training of normal delivery manoeuvres, we first recorded the hand trajectories of an experienced obstetrician interacting with a physical neonatal manikin using high-precision motion sensors. These trajectories were treated as expert demonstrations and subsequently used to generate animated virtual hand manoeuvres. During the MR simulation, the virtual expert hand model is rendered near the head of the physical neonatal manikin, whose spatial location is continuously tracked. A user interface component provides real-time visual prompts to guide the trainee's hands to spatially align with the expert hand posture, thereby facilitating skill acquisition through direct motion imitation.

%Figure 
\begin{figure}[h]
  % \vspace{-0.05in}
\centering
\includegraphics[width=0.43\textwidth]{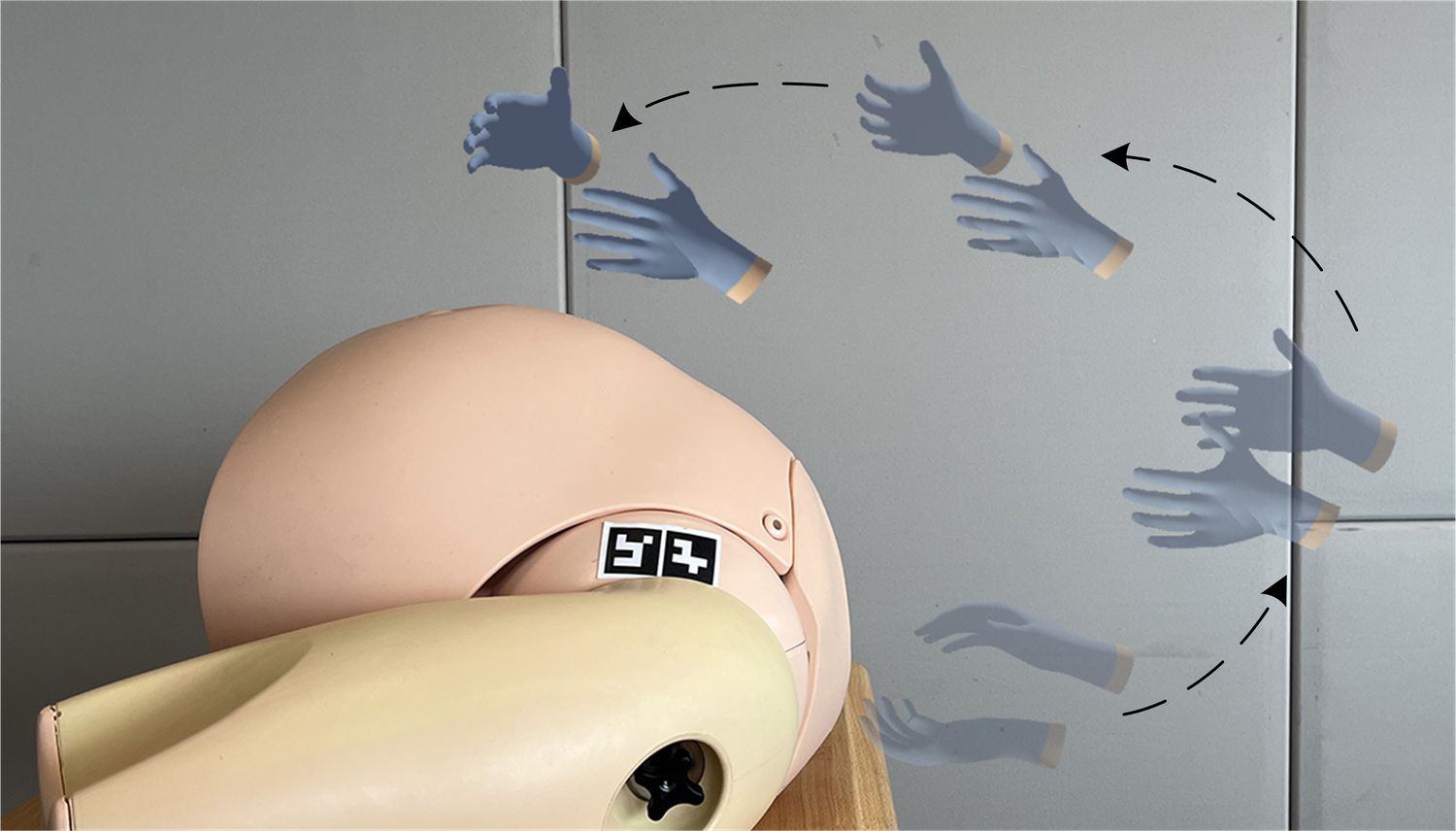}
\vspace{-0.05in}
\caption{Illustration of expert hand trajectory across the sequential manoeuvres of normal delivery under MR guidance.}
\label{fig:coorsystem}
\vspace{-0.05in}
\end{figure}

To account for practical variability in initial neonatal manikin poses, differing spatial arrangements between the trainee and the manikin, and other unpredictable training conditions, we designed a robust interaction mechanism that balances hand tracking precision and temporal smoothness. This design ensures consistent guidance quality and trainee comfort across diverse scenarios. A proximity-based trigger mechanism is implemented by attaching collision detectors to both the trainee’s wrist and the virtual expert hand. When the Euclidean distance between the two falls below 5 cm, the expert hand animation is activated. Multiple checkpoints are embedded along the animated trajectory to continuously monitor alignment accuracy. If the deviation between the trainee’s hand and the virtual trajectory exceeds a predefined threshold at any checkpoint, the animation is paused and a corrective prompt is issued, instructing the trainee to re-align with the expert hand before resuming.

This real-time hand guidance framework enables context-aware correction and promotes fine-grained motor skill acquisition, even under uncertain training conditions. By integrating adaptive animation control with spatial proximity sensing, the system provides an intelligent feedback loop that enhances both the fidelity and usability of MR-based procedural training.

% \vspace{0.10in}
\section{Evaluations}
\label{sec:Evaluations}

We evaluate our MR framework through both system-level registration quality analysis and trainee-level training assessment. The former examines calibration accuracy and model alignment, while the latter investigates usability and training outcomes.

% \vspace{0.10in}
\subsection{MR Registration Quality Assessment}
\label{subsec:MRAssessment}

\subsubsection{Assessment of Camera Calibration Accuracy}
\label{subsubsec:CalibrationAssessment}

\begin{figure}[h]
\vspace{-0.15in}
\centering
% \hspace{-0.02in}
\includegraphics[width=0.4\textwidth]{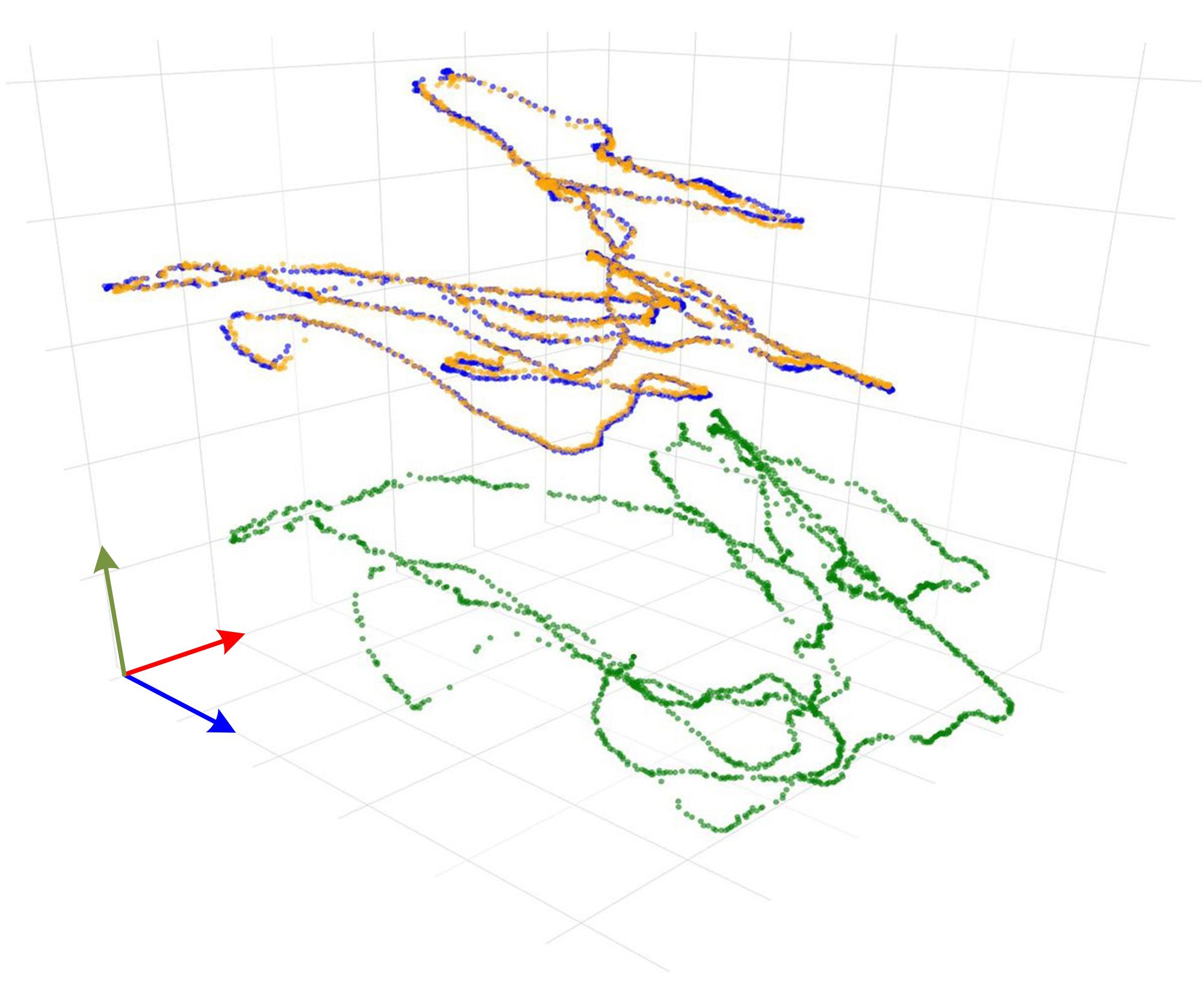}
\vspace{-0.3in}
\caption{Example assessment of camera calibration accuracy. 
\textcolor{green!60!black}{Green points} denote the original RGB-D camera trajectory, \textcolor{yellow!40!orange}{yellow points} denote the calibrated RGB-D camera trajectory and \textcolor{blue}{blue points} denote the Quest headset trajectory.}
\label{fig:CalibrationTest}
% \vspace{-0.05in}
\end{figure}

To evaluate the alignment accuracy between the RGB-D camera and Quest coordinate systems, a fiducial marker was rigidly attached to the center of the Quest controller's local coordinate system. The controller was then moved along arbitrary trajectories within the region visible to both the RGB-D camera and Quest headset. During this process, the raw movement trajectory of the fiducial marker is recorded by the RGB-D camera (green points in Fig. \ref{fig:CalibrationTest}) and the movement trajectory of the same point is captured by the Quest headset (blue points in Fig. \ref{fig:CalibrationTest}).

%Figure 2

Using the relative pose transformation $\mathbf{T}^{H}_{C}$ in Eq. (\ref{eq:eye2hand2}), the raw RGB-D trajectory was transformed into the corrected movement trajectory (yellow points in Fig. \ref{fig:CalibrationTest}). After $10$ independent trials, the RMSE between the corrected RGB-D trajectory and the Quest trajectory was calculated to be $3.6\pm 0.2 mm$. This calibration accuracy satisfies our requirements of childbirth training tasks and verifies the feasibility of hardwired connection between the Quest headset and RGB-D camera for MR task development.

% \vspace{-0.05in}
\begin{figure*}[t]
\centering
\subfloat[VR-Based Learning]{
  \hspace{-0.10in}
\includegraphics[height=0.278\textwidth]{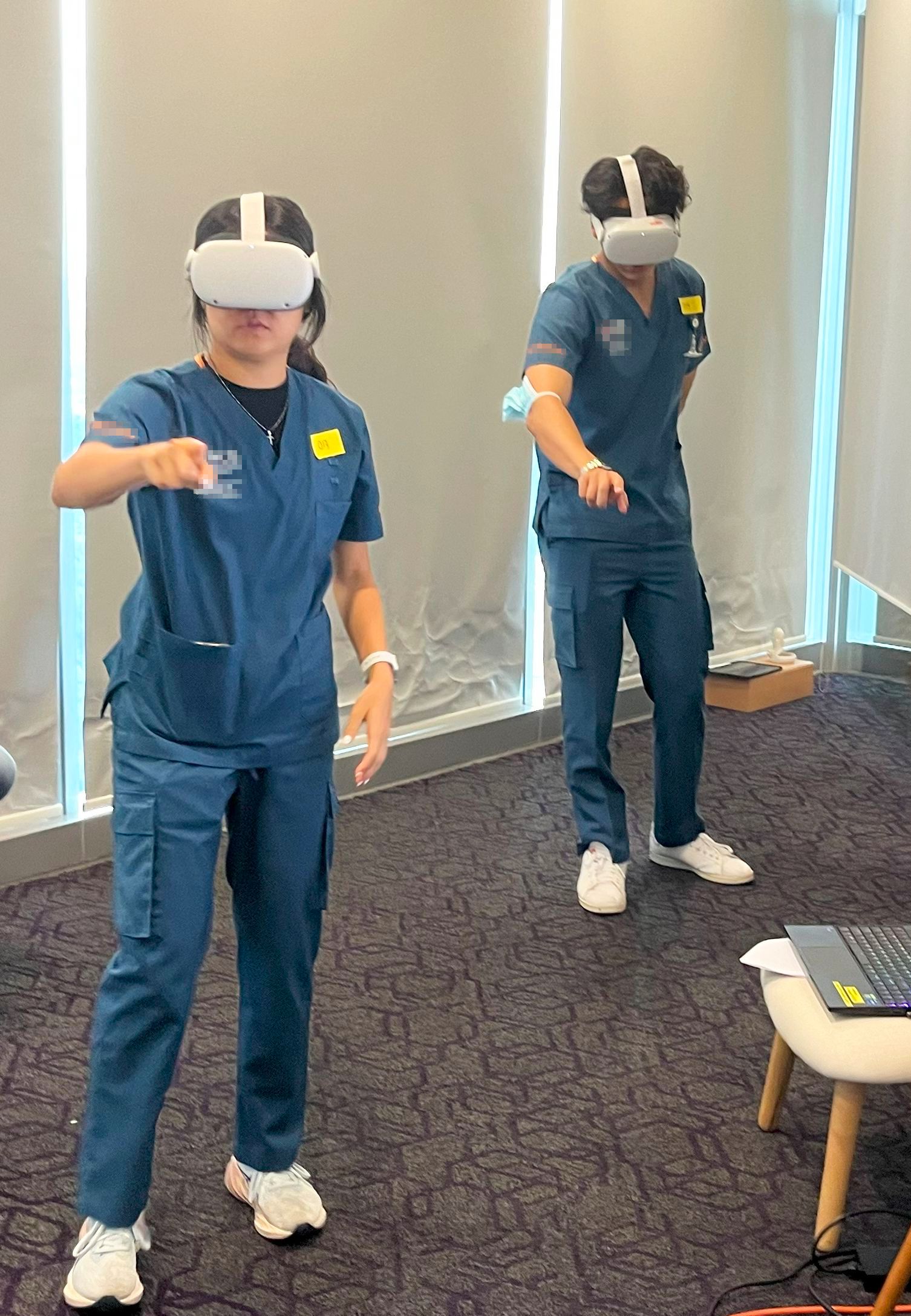} \label{subfig:VRus}
}
\subfloat[MR-Based Learning]{
  % \hspace{-0.05in}
\includegraphics[height=0.278\textwidth]{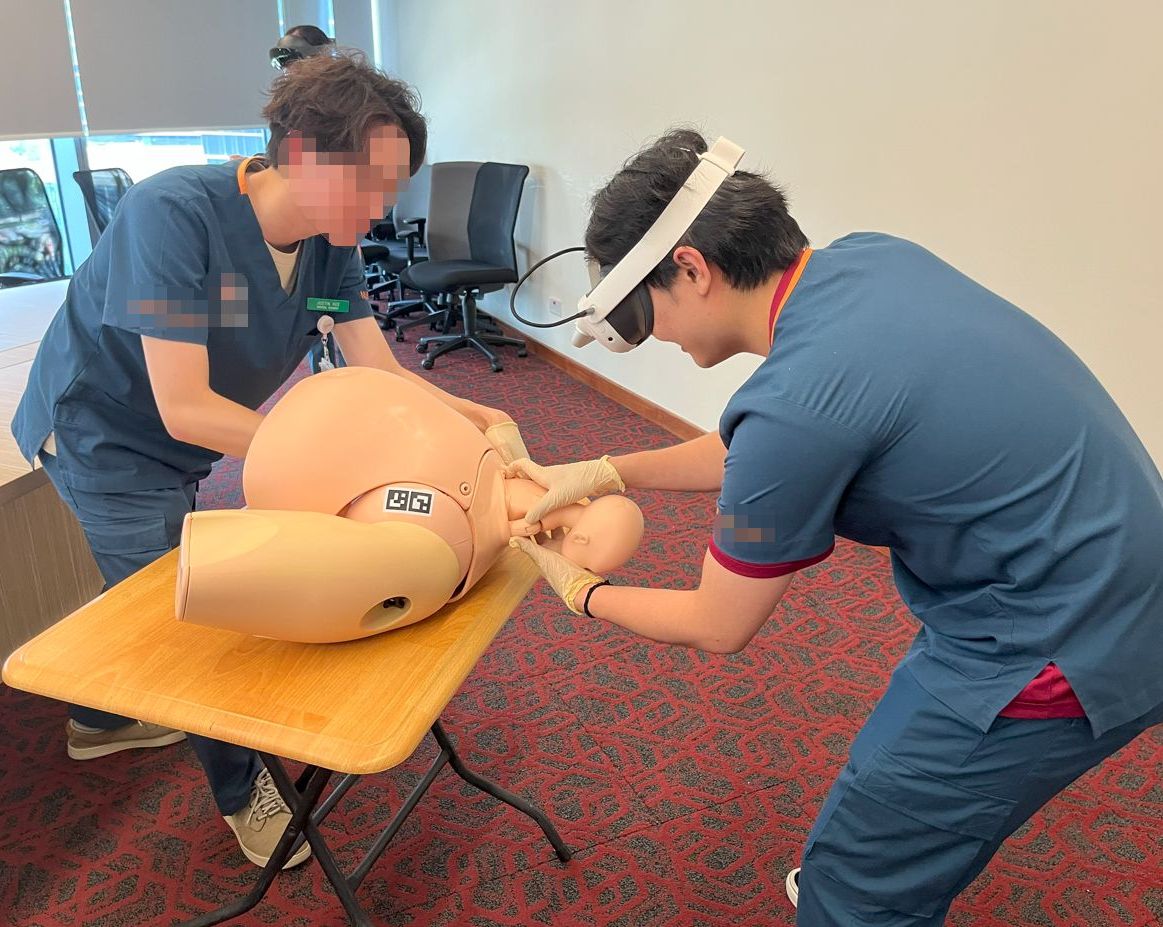} \label{subfig:MRus}}
\subfloat[DOPS Assessment]{
  % \hspace{-0.05in}
\includegraphics[height=0.278\textwidth]{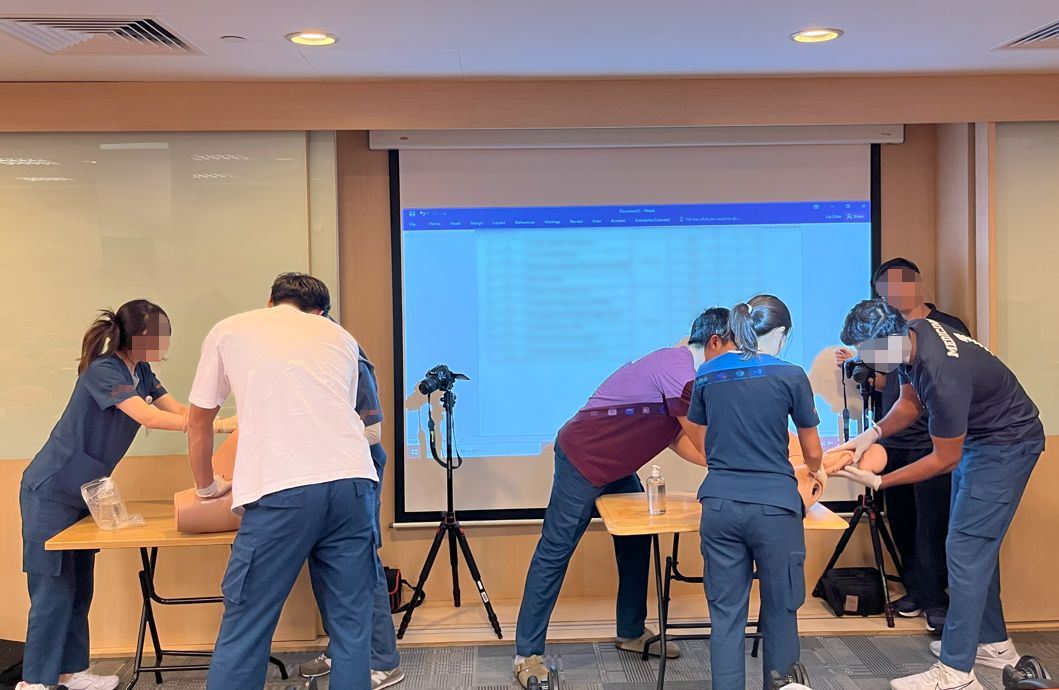} \label{subfig:DOPSus}}
\vspace{-0.05in}
\caption{Experimental setup of the user study. (a) Learning stage with VR-based childbirth training. (b) Learning stage with MR-based childbirth training integrating headset guidance with manikin interaction. (c) Assessment stage where both VR and MR groups performed the delivery task on a manikin without headset assistance, and were recorded by the video camera for the DOPS assessment.}
\label{fig:UECoordff}
% \vspace{-0.10in}
\end{figure*}

\subsubsection{Assessment of Manikin Alignment}
\label{subsubsec:MotherAlignmentAssessment}

To evaluate manikin tracking performance, we conducted a controlled experiment with the setups summarized in Tab.~\ref{tab:mothertracking}. During childbirth training, trainees primarily focus on the birthing region, and natural head movements can intermittently move fiducial markers out of the camera’s view. To address this and the limited surface area for marker placement, the manikin was instrumented with either two markers (one per side) or four markers (two per side). Given the manikin’s curved surface, depth data was also incorporated to examine whether it improves detection accuracy and stability. In practice, markers only need to be placed approximately symmetrically on the manikin surface to ensure visibility, without strict constraints on exact positions. The system does not assume a specific maternal manikin model either. When switching to a new manikin, the relative pose between the maternal manikin and the attached markers is computed once following the procedure described in Sec.~3.4.1. This calibration step is performed offline prior to training and does not involve trainee interaction.

In this experiment, only the maternal manikin alignment was assessed, as the neonatal head remains spatially constrained to the maternal structure. The delivery region estimated from the maternal alignment directly defines the subsequent neonatal head registration, which is performed only once per trial. Although the virtual expert hands are initially adjusted based on the neonatal head pose, their relative transformation to the maternal manikin remains fixed thereafter. Thus, accurate maternal alignment is the principal determinant of both the spatial fidelity of the delivery region and the reliability of neonatal head registration.

This experiment involves three key metrics, including Absolute Pose Error (APE), Jitter Frames Percentage (JFP) and Frames Per Second (FPS). APE evaluates global tracking accuracy by measuring errors in manikin translation and rotation. To obtain reliable ground truth for the APE metric, we rigidly attached a Quest Controller to the maternal manikin and used its recorded location and rotation as the reference pose. JFP assesses tracking stability through two sub-metrics. The first sub-metric, i.e., marker loss rate, accounts for the percentage of frames with undetected fiducial markers caused by occlusion or markers moving out of the camera's field of view. The second sub-metric, i.e., pose jittering rate, captures temporal inconsistencies in consecutive pose estimates caused by detection instability or noise. The pose jittering in our experiment is defined as a deviation exceeding $5 mm$ or $5^{\circ}$ between consecutive frames. FPS measures the tracking efficiency of maternal manikin on the Quest 3 headset's built-in processor. 

The experimental results in Tab. \ref{tab:mothertracking} show that increasing the number of markers leads to a reduction in APE, but only marginally in RGB-only setups. For instance, the shift from 2 to 4 markers in the RGB configuration results in a $4.4\%$ reduction in positional error and a $7.3\%$ reduction in rotational error. However, the inclusion of depth data substantially improves the tracking performance. Specifically, the 4-marker RGB-D configuration results in an $86\%$ reduction in positional error and an $80\%$ reduction in rotational error compared to the 4-marker RGB configuration, with a noticeable reduction in the standard deviation. This demonstrates that depth data is crucial in improving tracking accuracy, especially under real-world conditions where the manikin's curved surface could otherwise hinder marker detection.

\begin{table}[h]
  \footnotesize
\centering
\caption{Controlled experiment for tracking performance evaluation of the maternal manikin. ($\uparrow$ indicates higher values are better and $\downarrow$ indicates lower values are better.)}
\vspace{-0.05in}
\renewcommand{\arraystretch}{1.0}
\setlength\tabcolsep{1pt} 
\begin{tabular}{p{26mm}| C{18mm} C{18mm} C{17mm}}
\hlineB{3}
\centering \textbf{Setup} & \textbf{APE (mm/°) $\downarrow$} & \textbf{JFP (\%/\%) $\downarrow$} & \textbf{FPS $\uparrow$} \\
\hline
\hline
\multirow{2}{*}{2 Markers + RGB} & 25.34 ± 5.05 & 14.27 ± 2.82 & \multirow{2}{*}{\textbf{115.68 ± 5.19}} \\
& 19.25 ± 0.85 & 22.31 ± 3.08 & \\
\multirow{2}{*}{2 Markers + RGB-D} & 3.89 ± 0.36 & 14.13 ± 2.90 & \multirow{2}{*}{108.76 ± 4.22} \\
& 3.98 ± 0.32 & 6.98 ± 1.17 & \\

\multirow{2}{*}{4 Markers + RGB} & 24.22 ± 4.70 & 4.75 ± 1.58 & \multirow{2}{*}{110.29 ± 2.75} \\
& 17.84 ± 0.78 & 8.37 ± 2.67 & \\

\multirow{2}{*}{4 Markers + RGB-D} & \textbf{3.45 ± 0.26} & \textbf{4.71 ± 0.64} & \multirow{2}{*}{106.21 ± 2.23} \\
& \textbf{3.65 ± 0.28} & \textbf{2.96 ± 0.53} & \\
\hlineB{3}
\end{tabular}
% \vspace{-0.1in}
\label{tab:mothertracking}
\end{table}

For JFP, both depth data and marker count help improve tracking stability, but their effects differ between the two sub-metrics. The marker loss rate is primarily influenced by the number of markers, with 4-marker setups consistently reducing the loss rate by approximately $67\%$ compared to 2-marker setups. In contrast, the pose jittering rate benefits more significantly from the inclusion of depth data, with reductions of nearly $69\%$ when depth information is added to a 2-marker setup, and over $65\%$ in a 4-marker setup. These results highlight the substantial improvement in detection stability achieved through depth data, particularly on curved surfaces where RGB-only setups struggle to maintain consistent pose estimates. Finally, while the inclusion of depth data and additional markers slightly reduces FPS, the frame rate of 106 FPS in the 4-marker RGB-D setup remains comfortably within the real-time performance threshold, ensuring smooth and uninterrupted tracking during childbirth training scenarios.

% \vspace{0.10in}
\subsection{User Study Design}
\label{subsec:userstudy}

A large-scale user study was conducted in collaboration with the School of Medicine to evaluate immersive technologies for childbirth training by comparing MR and VR. The study protocol was reviewed and approved by the Institutional Review Board (IRB) of our university. The study focused on the normal delivery task and examined how physical interaction in MR may influence training outcomes relative to a fully virtual environment.

\subsubsection{Pipeline of User Study}
\label{subsubsec:Userstudypipeline}
The study was conducted during the first week of the Year 4 undergraduate medical students’ six-week O\&G clerkship and comprises two sessions: a \textit{learning session} on Monday and an \textit{assessment session} on Friday. 14 clinical groups, each consisting of six students, were recruited and cluster-randomised prior to the study.

\noindent{\textbf{Learning session.}} The learning session started off with participants watching a pre-recorded briefing by a medical expert in a meeting room. They were then distributed with a sticker of their assigned unique study record number (SRN) on, scanned the QR code and filled up their demographic information on their phone. Upon completion, they proceeded to a separate room for VR or MR simulations with on-site student helpers facilitating the session. The VR simulation (Fig.~\ref {subfig:VRus}) took on average 11 minutes, whereas the MR simulation (Fig.~\ref{subfig:MRus}) took slightly longer, averaging around 15 minutes, due to additional physical interaction and coordination with supporting trainees. Participants belonging to the MR group were additionally required to complete a preference questionnaire after finishing the MR simulation.

\noindent{\textbf{Assessment Session.}} On the day of the assessment session, participants gathered in a seminar room with 4 sets of Laerdal Prompt Flex manikins. All participants received surgical masks for anonymity and wrist tags indicating the SRN assigned to them. Following that, a medical expert demonstrated to them how to manipulate the manikin, focusing on inserting the neonatal manikin in and pushing it through the maternal manikin without hinting on the hand manoeuvre. After the demonstration, participants were grouped into teams of three, rotating through roles: expulsion of the neonate, stabilising the maternal manikin and neonate reception and handling. Participants then had five minutes to get familiar using the manikin. Four video cameras operated by study facilitators were set up to record the performance from fixed tripod positions focused on the manikin (Fig.~\ref{subfig:DOPSus}). Each participant was allowed two attempts to perform the hand manoeuvre assisting neonatal delivery. Before starting, they displayed the SRN on the wrist tag to the camera and then informed the facilitator by saying “I am ready”. Upon completing the procedure, participants indicated their completion by saying “I have completed”. After all recordings were captured, the session concluded, marking the end of the study.

\subsubsection{Assessment Tool}
\label{subsubsec:Participant_Questionnaire}

Two complementary tools were used: a \textit{Preference Questionnaire} for subjective perceptions of the MR system, and the \textit{Direct Observation of Procedural Skills} (DOPS) for expert evaluation of training performance in the assessment session.

\noindent{\textbf{Preference Questionnaire.}} The Preference Questionnaire was crafted based on Witmer and Singer’s Presence Questionnaire and the System Usability Scale. This concise questionnaire, designed to assess participants’ immediate perceptions of the MR system, contains five 5-point Likert-scale questions, with 1 indicating 'Not At All' and 5 indicating 'Completely'. The five rating questions were thoughtfully chosen to obtain comprehensive aspects of the MR simulation experience, and their key ideas can be extracted as listed in Table \ref{tab:mr_questionnaire}. As the questionnaire focuses on MR-specific interaction characteristics, it was administered only in the MR condition and not intended for a direct VR–MR comparison.

% The open-ended question is designed to capture the instantaneous feeling when the user transited from a purely virtual environment to a mixed environment. The user needs to use a word or a few words to simply describe that moment.

\noindent{\textbf{Direct Observation of Procedural Skills.}} 
DOPS was adapted as the primary assessment tool for the user study. It is a validated method in medical education used for evaluating medical students’ readiness, comprehension of procedures, and accuracy in technical skills. The technique grading rubric designed by the medical expert in the study team comprises four major categories, including hand placement prior to delivery, hand manoeuvre during delivery, handling of neonatal post-delivery, and overall carefulness during the procedure. (The detailed rubric is shown in the supplementary material.) Each sub-category is scored on a 3-point scale, with 2 points for correct independent execution, 1 point for inadequate performance, and 0 points for omission.

\begin{table}[h]
\footnotesize
\centering
\caption{Preference questionnaire used for MR evaluation. Each item was rated on a 5-point Likert scale.}
% \vspace{0.05in}
\renewcommand{\arraystretch}{1.0}
\setlength\tabcolsep{4pt}
\begin{tabular}{M{20mm}| L{58mm}}
\hlineB{3}
\centering \textbf{Dimension} & \multicolumn{1}{c}{\textbf{Description}}  \\
\hline
\hline
Engagement & How involved were you in the MR (passthrough) experience? \\ \hline
Presence & How well could you concentrate on the assigned tasks rather than the mechanisms used to perform them? \\ \hline
Ease of Use & How well could you move or manipulate objects in the MR environment? \\ \hline
Preference & I prefer the MR (passthrough) environment compared to the VR environment. \\ \hline
Interest in Reuse & I think I would like to practise using this simulation frequently. \\
\hlineB{3}
\end{tabular}
\label{tab:mr_questionnaire}
\end{table}

\vspace{-0.10in}
\begin{table}[h]
\footnotesize
\centering
\caption{Summary of assessment categories and sub-categories adapted from DOPS.}
% \vspace{0.05in}
\renewcommand{\arraystretch}{1.0}
\setlength\tabcolsep{6pt}
\begin{tabular}{M{20mm} | L{55mm}}
\hlineB{3}
\centering \textbf{Category} & \multicolumn{1}{c}{\textbf{Sub-category (3-point Scale)}} \\
\hline
\hline
\multirow{2}{*}{Hand Placement}
& Q1: Hand placement readiness \\
& Q2: Guarding of the perineum \\
\hline
\multirow{3}{*}{Delivery}
& Q3: Identification of extension \\
& Q4: Restitution with delivery of the anterior shoulder \\
\hline
\multirow{3}{*}{Post-Delivery}
& Q5: Confident handling of the neonate \\
& Q6: Placing the neonate on the abdomen of the maternal model \\
\hline
Overall 
& Q7: Overall handling and respect for tissue (any damage caused) \\
\hlineB{3}
\end{tabular}
\label{tab:dops_summary}
\end{table}

Four professional assessors from the O\&G Department, not affiliated with the study team, individually graded the video recordings of the performances. Prior to distribution for grading, videos were screened by the study team to exclude cases where the onsite instructor unintended intervention occurred or where camera obstruction blocked the visibility of key actions. To minimise grading bias potentially introduced by assessor fatigue or reduced attention span over time, all video recordings were randomly sequenced, ensuring a balanced mix of participants from different groups and assessment dates. Final scores were computed as the average among the ratings from four assessors.

\subsubsection{Demographics}
83 Year 4 undergraduate medical students from the Yong Loo Lin School of Medicine, National University of Singapore were recruited for this study. After accounting for absences, incomplete simulations and disqualified video recordings, 72 validated video recordings were included for grading, comprising 37 in the VR group and 35 in the MR group. The final cohort had an average age of 22.43 (SD = 0.90), with 52.78\% female participants and 47.22\% male participants. 98.61\% of the participants reported no prior experience with normal childbirth, with only one in the cohort having witnessed it once. Regarding VR usage, 72.22\% of the participants had used VR once a year, 26.39\% had never used VR, and only one reported relatively frequent usage of once a month.

\subsubsection{Results}
\label{subsubsec:SurveyFeedback}

\noindent{\textbf{Preference Questionnaire Statistics.}} 35 participants from the MR intervention group completed the Preference Questionnaire immediately after their MR simulation. The results of all five questions were not normally distributed based on the Shapiro-Wilk test. The \textit{Engagement} item received the highest mean score among the five items (Mean = 4.23, SD = 0.91). More than 80\% of the participants felt they were highly involved during the MR experience, highlighting MR’s capability in capturing learners’ attention and sustaining active participation. Moderately correlated with the \textit{Engagement} item (Spearman’s $\rho = 0.579$), the \textit{Presence} item received the second highest rating (Mean = 3.74, SD = 1.01). Despite potential distractions from their peers and the surroundings in the passthrough mixed reality mode, the participants still managed to concentrate on the given tasks and followed the hand manoeuvre throughout. The \textit{Ease of Use} item received a slightly lower score (Mean = 3.66, SD = 0.80) than \textit{Engagement} and \textit{Presence}. However, there was no major usability issue as none of the participants rated 1 and only two rated 2. 94.3\% of the participants did not encounter difficulty during their MR experience. 

Moreover, as the participants experienced both the fully virtual environment and the passthrough MR environment, the \textit{Preference} item was designed to determine whether they prefer training in a more isolated setting or a more interactive setting. The results showed that 5 participants preferred the former and 22 participants enjoyed the latter, while 8 remained neutral (Mean = 3.74, SD = 1.17). More vigorous discussions were observed in the MR group, driven by the inherently collaborative effort of manipulating the manikin. Lastly, the \textit{Interest in Reuse} item indicated that 20 participants (57.1\%) reported a strong willingness to use the designed MR system for future training (Mean = 3.51, SD = 1.17), reflecting its potential as a viable tool for skill acquisition and pointing to the need for further refinements to better align with medical students’ learning habits and practice workflows.

\begin{figure}[h]
\vspace{-0.05in}
\centering
% \hspace{-0.1in}
\includegraphics[width=0.47\textwidth]{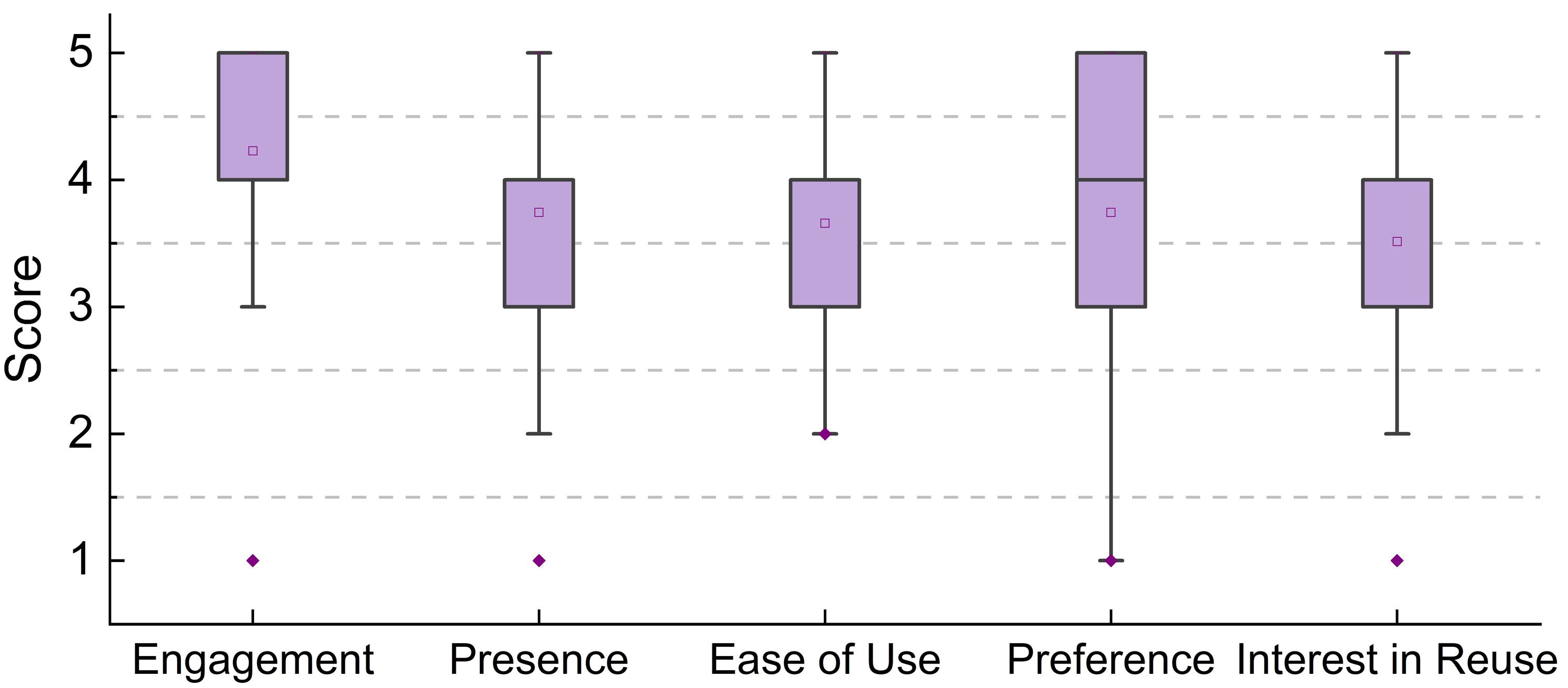}
% \vspace{-0.10in}
\caption{Statistical analysis of the preference questionnaire for MR evaluation.}
\label{fig:mrquestions}
\vspace{-0.05in}
\end{figure}

\begin{figure}[h]
% \vspace{-0.10in}
\centering
\hspace{-0.1in}
\includegraphics[width=0.47\textwidth]{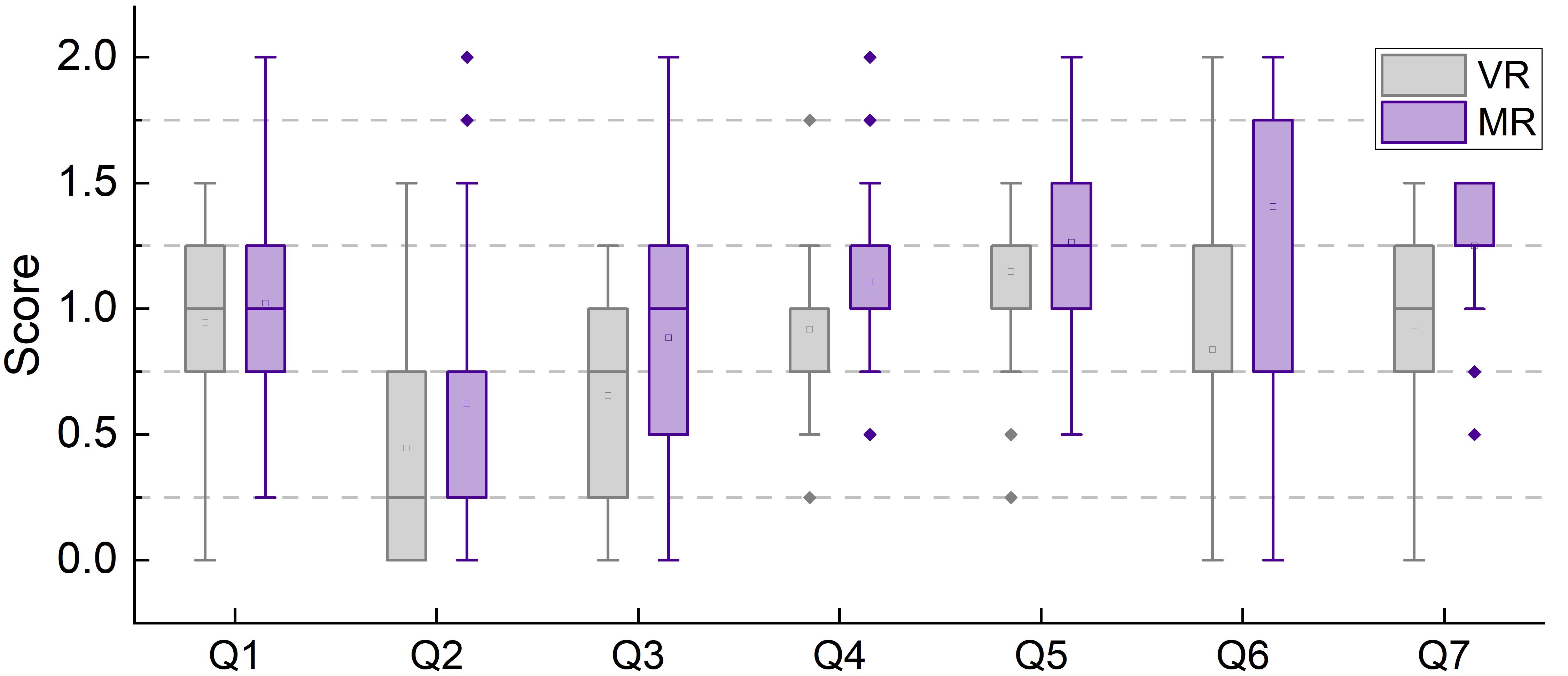}
% \vspace{-0.10in}
\caption{Statistical analysis of the DOPS across the categories in Table 3 comparing VR and MR conditions.}
\label{fig:dops}
\vspace{-0.05in}
\end{figure}

\noindent{\textbf{DOPS Statistics.}} Assumption of the same baseline performance was made since all participants were in their first week of their Year 4 O\&G clerkship and none had experience assisting an actual childbirth. The statistical distributions of the sub-category scores in Table \ref{tab:dops_summary} are visualized in Fig.\ref{fig:dops}. To enable group-level comparison, the scores of corresponding sub-categories were aggregated into composite scores. Since the Shapiro–Wilk test indicated that some of the aggregated scores were not normally distributed, a non-parametric Mann–Whitney test was used to evaluate pairwise differences between the VR and MR groups during the practical assessment. As shown in Fig.\ref{fig:dops} and Table \ref{tab:dops_stats}, \textit{Hand Placement} yielded comparable score distributions between VR and MR. This finding is consistent with expert interpretations that pre-delivery guidance did not differ substantially, as the physical manikin in MR offered minimal additional tactile input. In contrast, significant group differences were identified in \textit{Delivery}, \textit{Post-Delivery} and \textit{Overall Handling}. The consistently higher medians observed in MR, particularly in \textit{Post-Delivery}, suggest enhanced competence in critical tasks such as holding and positioning the newborn. Moreover, the narrower IQR in \textit{Overall Handling} reflects stronger inter-rater consensus, underscoring the reliability of expert evaluations and the greater proficiency conferred by MR training. 

Furthermore, the total scores across all categories in Table \ref{tab:dops_stats} were calculated to determine the overall extent to which the MR group outperformed the VR group. While the median of the total score of the VR group was $5.5$ (IQR = $3$), the MR group achieved a median of $7.5$ (IQR = $2$). The result indicated a statistically significant difference (p = $0.003$, U = $387$), demonstrating the MR group’s clear advantage over the VR group. Although the animation of the guiding hands is identical in both VR and MR simulations, instead of holding nothing in the air and simply following playback in VR, the compulsory action in MR, including feeling the tension of the neonate being delivered, sensing its weight in their hands, and eventually placing it back on the maternal manikin to complete the procedure. These requirements encourage sustained attentional engagement, as insufficient or imprecise manipulation makes it difficult for the neonate to be delivered. Consequently, the MR condition reinforces trainees’ focus on the hand manoeuvres and the procedural sequence, strengthening their memory of both the individual actions and the overall delivery process.

\begin{table}[h]
% \vspace{-0.05in}
\footnotesize
\centering
\caption{Statistical comparison of DOPS assessment scores using the Mann–Whitney test.}
\renewcommand{\arraystretch}{1.0}
\begin{tabular}{C{20mm}|C{13mm} C{14mm} C{6mm} C{12mm}}
\hlineB{3}
\makecell[c]{\textbf{Category}} & 
\makecell[c]{\textbf{Median} \\  \textbf{(VR / MR)}} & 
\makecell[c]{\textbf{IQR}\\ \textbf{(VR / MR)}} & 
\makecell[c]{\textbf{U}} & 
\makecell[c]{\textbf{p-value}} \\
\hline
\hline
Hand Placement & 1.5 / 1.5 & 1.00 / 1.00 & 552.5  & 0.284   \\
Delivery & 1.8 / 2.0  & 1.00 / 0.88 & 443.0  & p $<$ 0.05  \\
Post-Delivery & 2.0 / 2.8 & 1.25 / 1.25 & 327.0 & p $<$ 0.001 \\
Overall Handling & 1.0 / 1.3   & 0.50 / 0.25 & 365.5 & p $=$ 0.001 \\
\hlineB{3}
\end{tabular}
% \vspace{-0.05in}
\label{tab:dops_stats}
\end{table}

% Besides the objective ratings and subjective gradings, qualitative feedback was analyzed to capture participants’ first impressions of training in the MR environment. Beyond describing the experience as novel and engaging, many participants highlighted the added tactility of the physical manikin, which enhanced the sense of realism. Feedback also pointed to the need for refining the hand maneuver animation, as the manual handling of the neonatal manikin during MR training was more dynamic than the standardized neonatal motion animation in the VR simulation. Additional user feedback and detailed statistical results are provided in the supplementary material. 

To investigate whether participants’ subjective preference for MR training influenced their actual skill acquisition, we computed Spearman’s rank correlation between the rated preference total (Fig. \ref{fig:mrquestions}) and the graded DOPS total (Fig. \ref{fig:dops}). While the result suggested there was very weak correlation between the preference ratings and DOPS scores (Spearman’s $\rho = 0.051$), on the other hand, it suggested that participants’ personal preference towards the MR system did not influence their skill acquisition. It further implies that the design of this MR simulation is capable of effectively training the students, provided they follow the guided procedure, and has potential as an efficient tool for self-learning complicated hand manoeuvres.

% \vspace{-0.10in}
\section{Discussion and Future Work}
\label{sec:Discussion}

The effectiveness of the proposed MR system relies on the accuracy and temporal stability of the underlying spatial registration pipeline. In close-range manipulation scenarios such as childbirth delivery, even minor deviations can undermine the perceived alignment. Our findings from the user study indicate that the immersive interaction is driven by the fidelity of spatial coupling between visual guidance and physical interaction. By enabling trainees to directly map visual cues to physical contact locations without additional mental transformation, accurate spatial coupling constitutes a key system-level mechanism through which immersive interaction translates into effective skill acquisition. This perspective also helps clarify the limitations of prior works \cite{liu2024facilitating,ljungblad2025mixed} that rely on visually correct but spatially decoupled overlays, in which visual guidance is not spatially aligned with the physical contact space. Such systems implicitly assume that visual guidance alone is sufficient for skill acquisition, even when visual instructions and haptic interaction occur in misaligned reference frames. Our results suggest that this decoupling introduces spatial ambiguity during close-range manipulation, helping explain the limited transfer reported in VR-based training for tasks dominated by fine-grained hand–object interaction. Importantly, the proposed system does not require room-scale tracking infrastructure, which simplifies deployment and makes it more suitable for large-scale training scenarios.

Based on discussions with obstetric specialists, we identify several clinically motivated directions that are critical for bringing MR-based training closer to real delivery experience. First, the effectiveness of the proposed system is bounded by the assumption of rigid spatial registration. In normal childbirth scenarios, interaction is dominated by the global pose of the neonatal head, enabling rigid alignment to provide stable guidance. In contrast, complex cases such as breech presentation or shoulder dystocia violate this assumption due to non-rigid deformation and articulated motion, indicating the need for non-rigid neonatal pose estimation. Another implication concerns the use of pre-recorded expert hand trajectories. While this design ensures reproducibility and simplifies deployment, it assumes a fixed correspondence between expert demonstrations and the user’s physical interaction space. This assumption, though sufficient for the current study, may contribute to performance variability across users with different anthropometric characteristics, motivating anthropometry-aware guidance. Finally, MR-based training sessions reveal that accurate spatial alignment alone is insufficient to constrain safe manipulation. Some trainees applied excessive pulling forces, exposing a sim-to-real gap caused by the absence of force-aware feedback. This boundary of visual guidance suggests that spatial fidelity should be complemented by mechanisms that convey physical constraints, such as force feedback or stress visualization.

% \vspace{0.10in}
% \section{Limitations and Future Work}
% \label{sec:limitations}

% While our MR system demonstrates promising results in robust manikin localization for childbirth training, several limitations remain that suggest directions for future improvement. First, our current implementation focuses on normal childbirth training, which typically involves limited variation in hand manoeuvres. However, future work should extend the system to more complex delivery situations, such as breech presentation or shoulder dystocia, where precise localization of neonatal limbs and dynamic adaptation of obstetric techniques are critical. Addressing these scenarios would enhance the system's educational relevance and realism. Second, the current system relies on pre-recorded hand trajectories from expert clinicians to guide novice users. While this approach offers a structured demonstration of correct techniques, it lacks adaptability to individual learner variations such as body proportions (e.g., height and arm length). This may limit the personalization and intuitiveness of the training experience. To address this, future work will focus on developing adaptive guidance strategies that account for user-specific anthropometric differences, allowing for more fluid and natural interaction.

% \vspace{-0.2in}
\section{Conclusion}
\label{sec:Conclusions}

This research demonstrates the feasibility and pedagogical value of integrating MR into obstetric training. By aligning virtual models with physical manikins through a coarse-to-fine strategy, the system enables structurally faithful guidance that reinforces correct hand manoeuvres during delivery practice. Technical evaluations confirmed the accuracy, stability and real-time performance of the manikin localization pipeline on a standalone headset, validating the practicality of the approach. Importantly, a large-scale user study with medical students revealed that MR training not only enhanced task performance but was also consistently preferred to VR-based training. Taken together, these results highlight MR as a promising paradigm for simulation-based medical education. The proposed system reduces reliance on continuous expert supervision, offers scalable access to high-fidelity training, and provides a generalizable framework that could be extended to other manikin-based healthcare training scenarios.

\bibliographystyle{abbrv-doi}

% \clearpage
% \vspace{0.10in}
\bibliography{References}

%\vspace{-0.15in}

\end{document}